\documentclass[journal]{IEEEtran}

\usepackage{cite}
\usepackage{amsmath,amssymb}
\usepackage{graphicx}
\usepackage{booktabs}
\usepackage{tabularx}
\usepackage{multirow}
\usepackage{algorithm}
\usepackage{algpseudocode}
\usepackage{xcolor}
\usepackage[hidelinks]{hyperref}

\begin{document}

\title{Shared-Structure 4D Spectral Gaussian Representation for Sparse-View Spectral CT Reconstruction}

\author{Jiancheng~Fang,
        Shaoyu~Wang,
        Wenjun~Xia,~\IEEEmembership{Member,~IEEE},
        Yang~Chen,~\IEEEmembership{Senior Member,~IEEE},
        and Qiegen~Liu,~\IEEEmembership{Senior Member,~IEEE}%
\thanks{This work was supported in part by the National Natural Science Foundation of China under Grant 61871206 and Grant 62201628; in part by the National Key Research and Development Program of China under Grant 2022YFA1204200; and in part by the Guangdong Basic and Applied Basic Research Foundation under Grant 2023A1515011780. (Jiancheng Fang and Shaoyu Wang are co-first authors). (Corresponding author: Qiegen Liu).}%
\thanks{The acquisition of the mouse chest  data used in this study was approved by the Ethics Committee of the Li Ka Shing Faculty of Medicine, The University of Hong Kong, on February 20, 2023.}%
\thanks{Jiancheng Fang, Shaoyu Wang, and Qiegen Liu are with the School of Information Engineering, Nanchang University, Nanchang 330031, China 
(e-mail: d5283377@163.com; wangshaoyu@ncu.edu.cn; liuqiegen@ncu.edu.cn).}%
\thanks{Wenjun Xia and Yang Chen are with the School of Computer Science and Engineering, Southeast University, Nanjing 210096, China (e-mail: xwj90620@gmail.com; chenyang.list@seu.edu.cn). Yang Chen is also with the Jiangsu Provincial Joint International Research Laboratory of Medical Information Processing, Laboratory of Image Science and Technology, and Key Laboratory of New Generation Artificial Intelligence Technology and Its Interdisciplinary Applications (Southeast University), Ministry of Education.}%
}

\markboth{IEEE Transactions on Medical Imaging}%
{Fang \MakeLowercase{\textit{et al.}}: Shared-Structure 4D Spectral Gaussian Representation}

\maketitle

\begin{abstract}
Sparse-view spectral computed tomography (CT) reconstructs energy-resolved attenuation volumes from limited projection views, requiring simultaneous handling of angular undersampling and spectral coupling. We propose a Shared-Structure 4D Spectral Gaussian Representation (4D-SG) that learns shared Gaussian geometry from full spectrum structural projections and uses a Gaussian-wise Spectral Density Curve Network (GSC-Net) to predict Gaussian raw density transformations. This factorization separates shared spatial structure from spectral attenuation variation, avoids independent channel geometry optimization, and establishes a continuous 4D-SG representation from discrete spectral measurements for unobserved spectral channel queries. Experiments on six synthesized, simulated projection, and real projection datasets with 50 views demonstrate the best average performance. Compared with the strongest Gaussian baseline, 4D-SG improves PSNR from 35.56 dB to 36.61 dB, increases SSIM from 0.909 to 0.914, and reduces LPIPS from 0.208 to 0.194, demonstrating its effectiveness for sparse-view spectral CT reconstruction.
\end{abstract}

\begin{IEEEkeywords}
Sparse-view reconstruction, spectral CT, Gaussian splatting, spectral modeling.
\end{IEEEkeywords}

\section{Introduction}

\begin{figure}[t]
    \centering
    \includegraphics[width=\columnwidth]{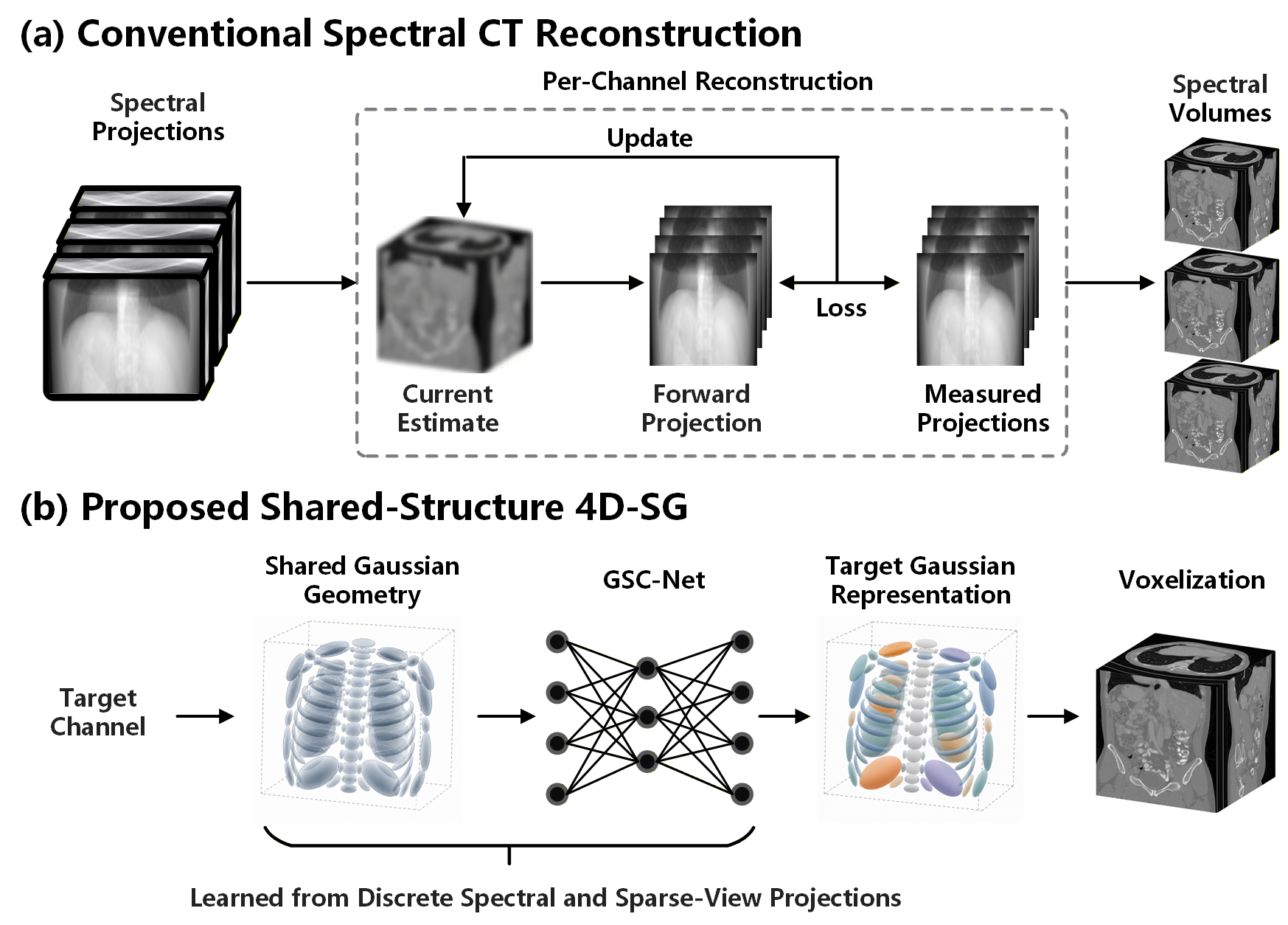}
    \caption{Conceptual comparison. (a) Conventional methods reconstruct each spectral channel through an independent iterative process. (b) 4D-SG learns shared Gaussian geometry and GSC-Net from discrete spectral measurements and sparse-view projections and generates the Gaussian representation of a target channel for voxelization, supporting reconstruction and prediction of observed and unobserved spectral channels.}
    \label{fig:intro}
\end{figure}

\IEEEPARstart{S}{pectral} computed tomography (CT), including dual-energy CT, multi-energy CT, and photon-counting CT, provides energy-resolved measurements beyond conventional attenuation-based CT \cite{mccollough2015dual,willemink2018photon,greffier2023spectral}. Its energy-dependent attenuation enables multi-energy attenuation imaging, material discrimination, virtual monoenergetic imaging, and other quantitative spectral imaging applications. Reconstruction requires multiple correlated attenuation volumes with different spectral responses, making spectral modeling across channels essential.

Sparse-view acquisition reduces radiation dose, scan time, and system complexity in spectral CT. However, angular undersampling and spectral channel separation occur simultaneously: each channel contains incomplete projection information, while all channels correspond to the same object. Reconstruction must therefore suppress sparse-view artifacts, preserve structural consistency across channels, and retain attenuation differences between spectral channels \cite{rigie2015joint,zhang2017tensor,liu2024sgm}.

Existing spectral CT methods exploit physical forward models, correlations across channels, or learned image priors. Model-based approaches include statistical reconstruction, joint regularization, tensor representations, basis-image estimation, and calibrated spectral forward models \cite{long2014multimaterial,rigie2015joint,zhang2017tensor,schmidt2017basis,mechlem2018joint,shen2023josr}, while learning-based approaches use deep spectral imaging pipelines, model-based unrolling, learned structural priors, and generative constraints \cite{wu2021dlspectral,chen2024soulnet,shi2024textureprior,liu2024sgm,guo2025coupled,zhang2024oign}. Most of these methods still represent spectral CT volumes as voxel grids, image tensors, or separate channel outputs, coupling shared spatial structure with spectral attenuation variation and limiting controlled reuse of common 3D structure.

Representation learning offers an alternative for sparse-view tomographic reconstruction. Implicit neural representations model continuous attenuation fields as functions of spatial coordinates and have been applied to sparse-view CT, cone-beam CT, and X-ray 3D reconstruction \cite{song2023piner,lin2023difnet,lin2024c2rv,cai2024saxnerf}. More recently, 3D Gaussian Splatting (3DGS) has provided compact learnable Gaussian primitives and efficient differentiable rendering \cite{kerbl2023gaussian}, and Gaussian representations have been extended to X-ray imaging and CT reconstruction \cite{cai2024xgaussian,zha2024r2gaussian,lin2024learning,li20253dgrct}. However, existing methods mainly address generic tomography, cone-beam CT, or single-energy sparse-view CT rather than shared object geometry with varying spectral attenuation responses.

We propose a Shared-Structure 4D Spectral Gaussian Representation (4D-SG) for sparse-view spectral CT reconstruction. Instead of assigning an independent volumetric representation to each spectral channel, 4D-SG learns shared Gaussian geometry for common 3D structure and Gaussian raw density curves for spectral attenuation variation, as illustrated in Fig.~\ref{fig:intro}.

The main contributions are summarized as follows:
\begin{itemize}
    \item We introduce, to the best of our knowledge, the first Gaussian-splatting framework for 4D spectral CT reconstruction; 4D-SG separates shared Gaussian geometry from attenuation responses across spectral channels.

    \item We develop a \emph{Gaussian-wise Spectral Density Curve Network} (GSC-Net) to predict raw density transformations over the ordered spectral axis, using channel correlations while preserving channel-specific attenuation differences.

    \item We establish a continuous 4D-SG representation over the spectral axis from discrete measurements, supporting reconstruction conditioned on the spectral coordinate and queries of unobserved channels for flexible recovery of multi-energy attenuation volumes.
\end{itemize}

\section{Related Work}

\begin{figure}[t]
    \centering
    \includegraphics[width=\columnwidth]{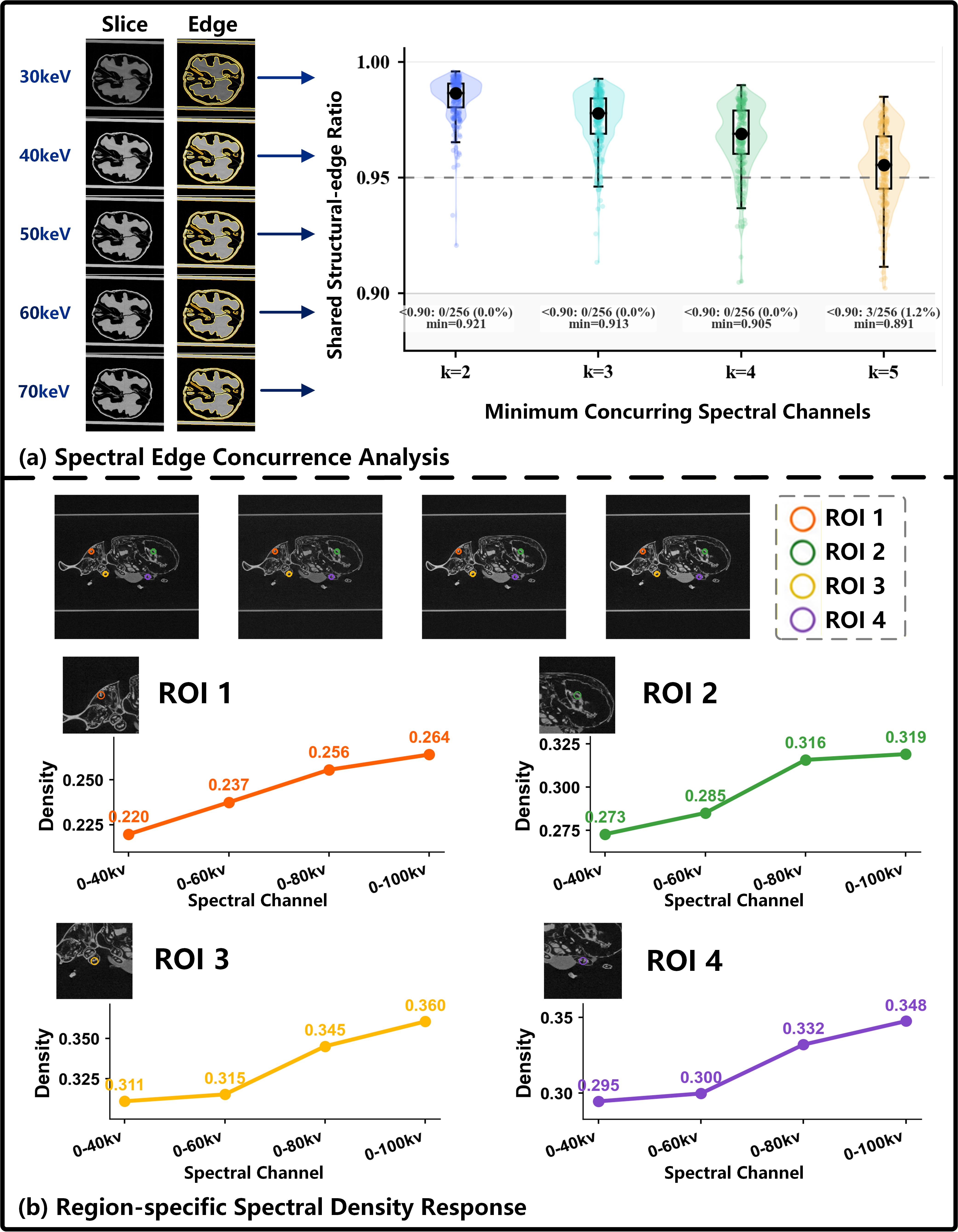}
    \caption{Empirical analysis of structural consistency across channels and attenuation variation by region in spectral CT. (a) Representative spectral CT slices, corresponding edge maps, and distributions of the shared structural edge ratio under different thresholds $k$ for the minimum number of concurring channels. Violin plots, samples from individual slices, box plots, and median markers summarize the distributions; the dashed line denotes 0.95, and the bottom annotations report the proportion of samples with low ratios and the minimum ratio. (b) Four fixed regions of interest, enlarged patches, and their mean globally normalized attenuation values across spectral channels, with the measured values annotated at each channel.}
    \label{fig:motivation}
\end{figure}

\subsection{Spectral CT Reconstruction}

Existing spectral CT reconstruction methods can be broadly categorized into model-based and learning-based approaches. Spectral CT, including dual-energy, multi-energy, and photon-counting CT, reconstructs multiple energy-dependent attenuation images, making channel correlation particularly important under sparse-view acquisition \cite{mccollough2015dual,greffier2023spectral,wang2021spectralpcct}.

Model-based approaches include statistical image reconstruction for material decomposition \cite{long2014multimaterial}, joint multi-channel or total-image-constrained regularization \cite{rigie2015joint,liu2016ticmr}, tensor-based dictionary learning \cite{zhang2017tensor}, basis-image estimation \cite{schmidt2017basis}, and joint statistical iterative reconstruction with calibrated forward models \cite{mechlem2018joint}. More recent methods jointly estimate image content and spectral response parameters \cite{shen2023josr}. Although physically interpretable and effective at exploiting spectral coupling, these approaches require repeated projection and backprojection over multiple channels and can be computationally costly for 3D sparse-view reconstruction.

Learning-based spectral CT reconstruction methods \cite{wu2021dlspectral,bousse2024review} include deep spectral imaging pipelines and challenge-driven reconstruction \cite{hu2024prospect,sidky2024challenge}, model-based unrolling \cite{ge2023mbdectnet,chen2024soulnet}, learned texture or structural priors \cite{shi2024textureprior,liu2025spn}, and generative or structural constraints for sparse-view spectral reconstruction \cite{liu2024sgm,guo2025coupled,zhang2024oign}. They improve artifact suppression and detail recovery but generally rely on voxel grids, image tensors, or separate spectral-channel outputs, coupling shared spatial structure with spectral attenuation variation.

\subsection{Implicit Neural and Gaussian Representations}

Implicit neural representations model attenuation fields as continuous spatial functions and have been applied to sparse-view and cone-beam CT reconstruction \cite{song2023piner,lin2023difnet,lin2024c2rv,cai2024saxnerf}. Despite reducing dependence on voxel grids, they require dense coordinate sampling and repeated multilayer perceptron evaluations, limiting reconstruction and rendering efficiency.

\begin{figure*}[t]
    \centering
    \includegraphics[width=1.0\textwidth]{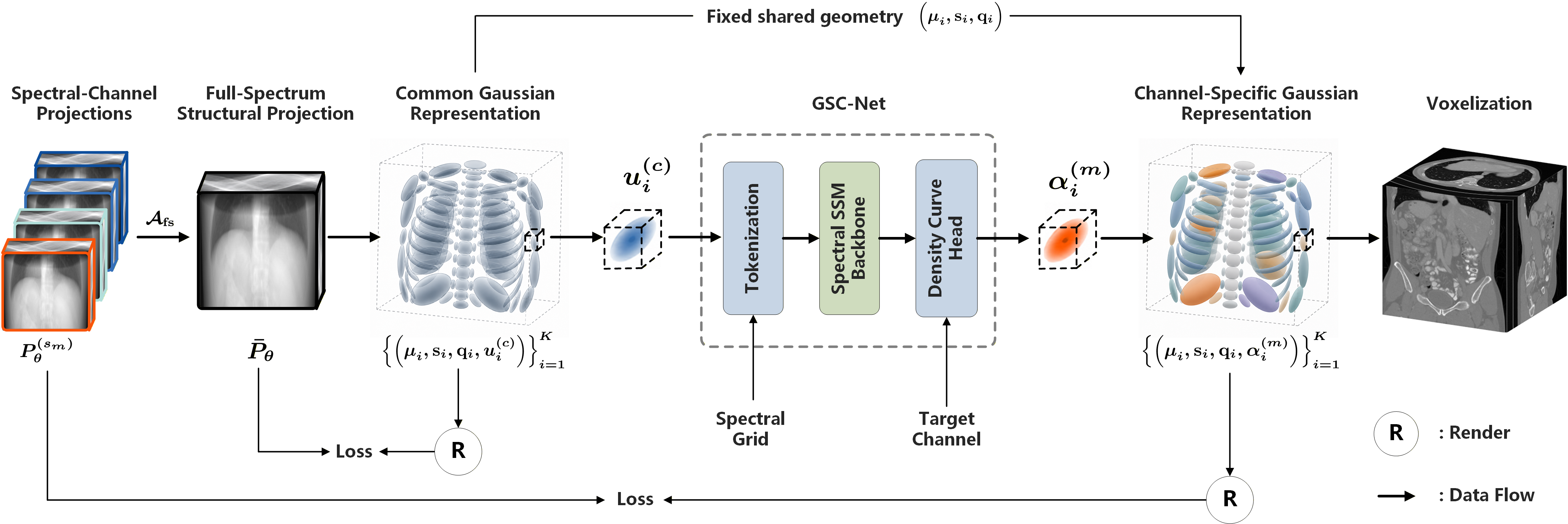}
    \caption{Workflow of 4D-SG. Shared Gaussian geometry and common raw densities are first learned from structural projections over the full spectrum using projection fidelity and volumetric regularization. With the Gaussian centers, scales, and rotations fixed, GSC-Net predicts raw density transformations for each Gaussian conditioned on the common raw densities, spectral grid, and target channel. The resulting density coefficients for the target channel are combined with the fixed shared geometry and optimized using spectral projection fidelity, curve regularization, and volumetric regularization. Gaussian voxelization reconstructs the queried spectral channel, supporting reconstruction of observed spectral channels and queries of unobserved spectral channels.}
    \label{fig:framework}
\end{figure*}

3DGS represents geometry and attributes using learnable Gaussian primitives, providing a compact parameterization and efficient differentiable rendering \cite{kerbl2023gaussian}. Gaussian representations have been extended to X-ray novel view synthesis, tomographic reconstruction, extremely sparse-view cone-beam CT reconstruction, and sparse-view CT reconstruction \cite{cai2024xgaussian,zha2024r2gaussian,lin2024learning,li20253dgrct}. However, these methods mainly target generic tomography, cone-beam CT, or single-energy sparse-view CT. Separation of shared Gaussian geometry and spectral attenuation variation remains unaddressed in spectral CT.

\section{Method}

\subsection{Motivation}

Spectral CT channels depict the same object and therefore share a common spatial support, whereas the attenuation values vary as the spectral coordinate changes. Fig.~\ref{fig:motivation} provides empirical evidence for this distinction between shared spatial structure and spectral attenuation variation.

As shown in Fig.~\ref{fig:motivation}(a), spectral slices preserve consistent boundaries, and the shared structural edge ratio remains high across different thresholds $k$ for the minimum number of concurring channels. This consistency supports reusing Gaussian geometry across channels. Independent optimization introduces redundant parameters and may cause structural drift between channels. Meanwhile, Fig.~\ref{fig:motivation}(b) shows that the mean globally normalized attenuation of four fixed regions of interest (ROIs) follows systematic, ordered changes across spectral channels. This phenomenon supports modeling attenuation as a correlated function along the spectral axis rather than as unrelated values across channels.

Based on the observations, 4D-SG factorizes the spectral CT representation into shared Gaussian geometry and raw density transformation curves for each Gaussian. The shared Gaussian geometry captures the 3D structure common to all spectral channels, while the transformation curves describe spatially varying attenuation as it changes along the spectral axis.

\begin{figure*}[t]
    \centering
    \includegraphics[width=0.88\textwidth]{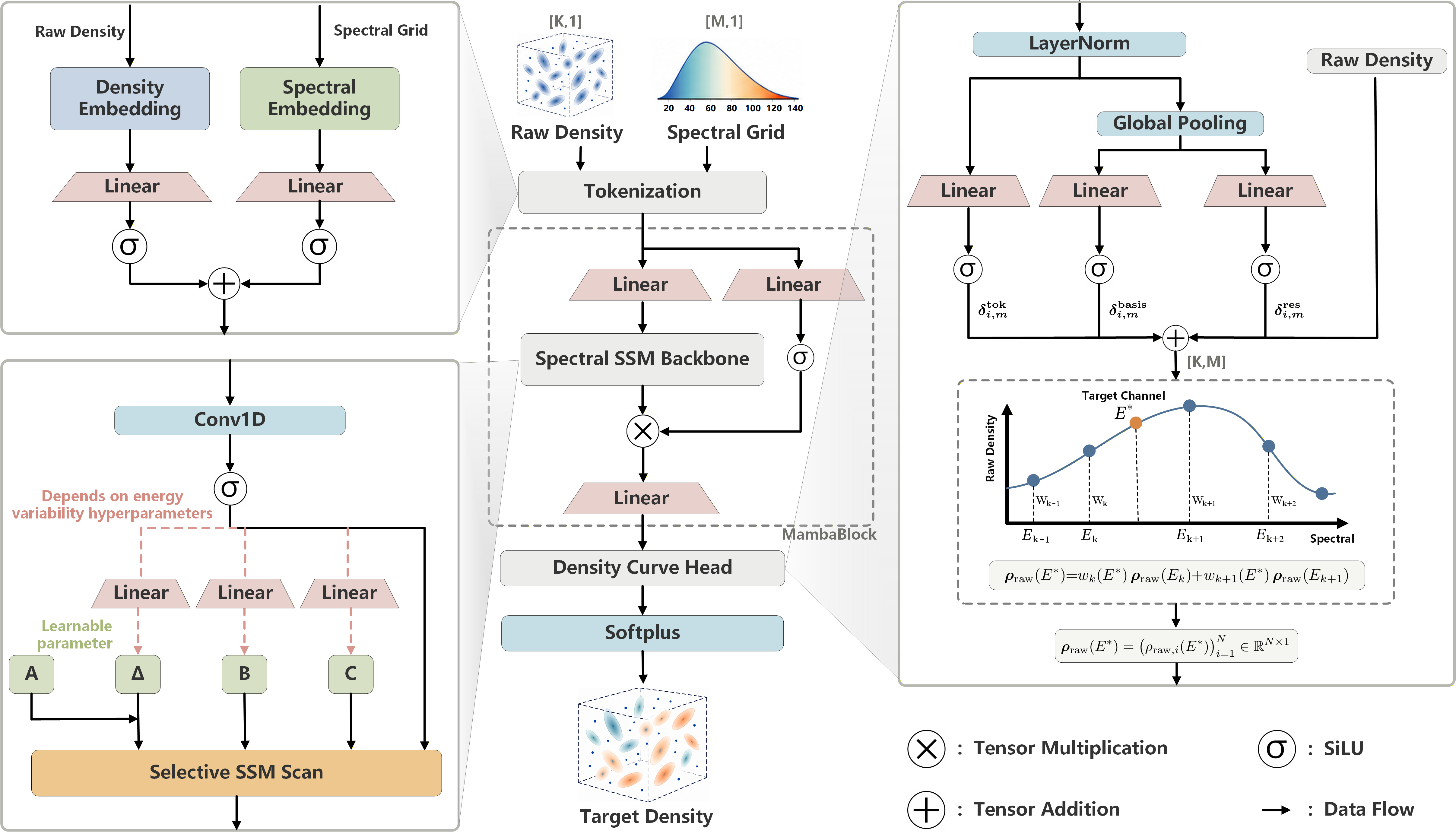}
    \caption{Architecture of GSC-Net. The common raw densities and normalized spectral grid are embedded into spectral token sequences for each Gaussian and processed by the gated spectral SSM backbone. The density curve head combines token, low-rank basis, and residual responses to predict raw density transformations for each Gaussian. Observed spectral channels are queried directly from the predicted curves, whereas target spectral channels are queried from the predicted raw density curves before conversion to non-negative density coefficients.}
    \label{fig:net}
\end{figure*}

\subsection{Theoretical Rationale}
\label{sec:theoretical_rationale}

The structural consistency and ordered attenuation variation in
Fig.~\ref{fig:motivation} motivates separating shared spatial structure
from spectral attenuation variation. With $s$ denoting a generic spectral coordinate, a material basis model represents the attenuation field as
$\mu(\mathbf{x},s)=\sum_{b=1}^{B}\rho_b(\mathbf{x})f_b(s)$,
where $\rho_b(\mathbf{x})$ and $f_b(s)$ are the spatial distribution and
spectral response of the $b$-th basis component, respectively. The basis
distributions are approximated using a shared set of Gaussian primitives:
\begin{equation}
\rho_b(\mathbf{x})
\approx
\sum_{i=1}^{K}
\beta_{i,b}
G_i(\mathbf{x};
\boldsymbol{\mu}_i,\mathbf{\Sigma}_i).
\label{eq:material_gaussian_factorization}
\end{equation}
The Gaussian kernel and covariance matrix are defined as
\begin{equation}
\resizebox{0.95\columnwidth}{!}{$\displaystyle G_i(\mathbf{x};\boldsymbol{\mu}_i,\mathbf{\Sigma}_i)=\exp\left[-\frac{1}{2}(\mathbf{x}-\boldsymbol{\mu}_i)^{\top}\mathbf{\Sigma}_i^{-1}(\mathbf{x}-\boldsymbol{\mu}_i)\right],\qquad \mathbf{\Sigma}_i=\mathbf{R}(\mathbf{q}_i)\operatorname{diag}(\mathbf{s}_i^2)\mathbf{R}(\mathbf{q}_i)^{\top}.$}
\label{eq:gaussian_kernel_covariance}
\end{equation}
Thus,
$\{(\boldsymbol{\mu}_i,\mathbf{s}_i,\mathbf{q}_i)\}_{i=1}^{K}$
defines the shared Gaussian geometry. We write
$G_i(\mathbf{x})=
G_i(\mathbf{x};\boldsymbol{\mu}_i,\mathbf{\Sigma}_i)$
when the geometry is fixed. Substituting
Eq.~\eqref{eq:material_gaussian_factorization} into the material basis
model gives
\begin{equation}
\hat{\mu}(\mathbf{x},s)
=
\sum_{b=1}^{B}
\sum_{i=1}^{K}
\beta_{i,b}f_b(s)G_i(\mathbf{x}).
\label{eq:substitute_gaussian_basis}
\end{equation}
With 
$\alpha_i(s)=\sum_{b=1}^{B}\beta_{i,b}f_b(s)$, the representation is written as follows:
\begin{equation}
\hat{\mu}(\mathbf{x},s)
=
\sum_{i=1}^{K}
\alpha_i(s)G_i(\mathbf{x}).
\label{eq:continuous_4d_sg_representation}
\end{equation}
This representation separates shared Gaussian geometry from spectral
density coefficients.
To examine geometry sharing, we consider a general representation in which
both density coefficients and Gaussian geometry vary with $s$:
\begin{equation}
\widetilde{\mu}(\mathbf{x},s)
=
\sum_{i=1}^{K}
\alpha_i(s)
G_i\left(
\mathbf{x};
\boldsymbol{\mu}_i(s),
\mathbf{\Sigma}_i(s)
\right).
\label{eq:varying_geometry_representation}
\end{equation}
Define
$\mathbf{r}_i(s)=\mathbf{x}-\boldsymbol{\mu}_i(s)$,
$\mathbf{M}_i(s)=\mathbf{\Sigma}_i(s)^{-1}$, and
$G_i(\mathbf{x},s)=
G_i(\mathbf{x};
\boldsymbol{\mu}_i(s),\mathbf{\Sigma}_i(s))$.
Let
$\phi_i(s)=
-\frac{1}{2}\mathbf{r}_i(s)^{\top}
\mathbf{M}_i(s)\mathbf{r}_i(s)$.
The argument $s$ is omitted below when unambiguous. Applying the matrix
product rule gives
\begin{equation}
\phi_i'=-\frac{1}{2}\Big[\mathbf{r}_i'^{\top}\mathbf{M}_i\mathbf{r}_i+\mathbf{r}_i^{\top}\mathbf{M}_i'\mathbf{r}_i+\mathbf{r}_i^{\top}\mathbf{M}_i\mathbf{r}_i'\Big].
\label{eq:gaussian_exponent_product_rule}
\end{equation}
Using
$\mathbf{r}_i'=-\boldsymbol{\mu}_i'$ and
$\mathbf{M}_i'=
-\mathbf{M}_i\mathbf{\Sigma}_i'\mathbf{M}_i$,
Eq.~\eqref{eq:gaussian_exponent_product_rule} becomes
\begin{equation}
\phi_i'=\boldsymbol{\mu}_i'^{\top}\mathbf{M}_i\mathbf{r}_i+\frac{1}{2}\mathbf{r}_i^{\top}\mathbf{M}_i\mathbf{\Sigma}_i'\mathbf{M}_i\mathbf{r}_i.
\label{eq:gaussian_exponent_derivative}
\end{equation}
Using the symmetry of $\mathbf{M}_i$ and
$G_i(\mathbf{x},s)=\exp(\phi_i(s))$, the chain rule gives
\begin{equation}
\resizebox{0.95\columnwidth}{!}{$\displaystyle \frac{\partial G_i(\mathbf{x},s)}{\partial s}=G_i(\mathbf{x},s)\boldsymbol{\mu}_i'^{\top}\mathbf{M}_i\mathbf{r}_i+\frac{1}{2}G_i(\mathbf{x},s)\mathbf{r}_i^{\top}\mathbf{M}_i\mathbf{\Sigma}_i'\mathbf{M}_i\mathbf{r}_i.$}
\label{eq:gaussian_kernel_derivative}
\end{equation}

With
$\mathbf{R}_i(s)=\mathbf{R}(\mathbf{q}_i(s))$ and
$\mathbf{D}_i(s)=\operatorname{diag}(\mathbf{s}_i(s)^2)$.
from
$\mathbf{\Sigma}_i(s)=
\mathbf{R}_i(s)\mathbf{D}_i(s)\mathbf{R}_i(s)^{\top}$,
we obtain
\begin{equation}
\mathbf{\Sigma}_i'=\mathbf{R}_i'\mathbf{D}_i\mathbf{R}_i^{\top}+\mathbf{R}_i\mathbf{D}_i'\mathbf{R}_i^{\top}+\mathbf{R}_i\mathbf{D}_i\mathbf{R}_i'^{\top},
\label{eq:covariance_derivative}
\end{equation}
where
$\mathbf{D}_i'
=
2\operatorname{diag}
(\mathbf{s}_i\odot\mathbf{s}_i')$.
Differentiating Eq.~\eqref{eq:varying_geometry_representation} and
substituting Eq.~\eqref{eq:gaussian_kernel_derivative} yields
\begin{equation}
\resizebox{0.95\columnwidth}{!}{$\displaystyle
\begin{aligned}
\frac{\partial\widetilde{\mu}(\mathbf{x},s)}{\partial s}
={}&
\sum_{i=1}^{K}\alpha_i'G_i(\mathbf{x},s)
+
\sum_{i=1}^{K}\alpha_iG_i(\mathbf{x},s)
\boldsymbol{\mu}_i'^{\top}\mathbf{M}_i\mathbf{r}_i \\
&+
\frac{1}{2}\sum_{i=1}^{K}\alpha_iG_i(\mathbf{x},s)
\mathbf{r}_i^{\top}\mathbf{M}_i
\mathbf{\Sigma}_i'\mathbf{M}_i\mathbf{r}_i .
\end{aligned}
$}
\label{eq:spectral_variation_decomposition}
\end{equation}
The three terms represent density variation, Gaussian center
displacement, and covariance deformation, respectively.

In 4D-SG, the Gaussian centers, scales, and rotations are independent of
$s$. The last two terms in
Eq.~\eqref{eq:spectral_variation_decomposition} therefore vanish:
\begin{equation}
\frac{\partial\hat{\mu}(\mathbf{x},s)}
{\partial s}
=
\sum_{i=1}^{K}
\alpha_i'(s)G_i(\mathbf{x}).
\label{eq:shared_geometry_spectral_derivative}
\end{equation}
Thus, spectral variation is represented only by Gaussian density changes
over fixed spatial primitives. Integrating
Eq.~\eqref{eq:shared_geometry_spectral_derivative} over $[s_a,s_b]$
gives
\begin{equation}
\resizebox{0.95\columnwidth}{!}{$\displaystyle \hat{\mu}(\mathbf{x},s_b)-\hat{\mu}(\mathbf{x},s_a)=\int_{s_a}^{s_b}\frac{\partial\hat{\mu}(\mathbf{x},s)}{\partial s}\,\mathrm{d}s=\sum_{i=1}^{K}\left[\alpha_i(s_b)-\alpha_i(s_a)\right]G_i(\mathbf{x}).$}
\label{eq:cross_spectral_difference}
\end{equation}
Hence, any two spectral channels remain represented by the same Gaussian
primitives.

For spectral weights $\{w_m\}_{m=1}^{M}$, linear spectral aggregation is
defined as
\begin{equation}
\bar{\mu}(\mathbf{x})
=
\sum_{m=1}^{M}
w_m\hat{\mu}(\mathbf{x},s_m).
\label{eq:spectral_aggregation_definition}
\end{equation}
Substituting Eq.~\eqref{eq:continuous_4d_sg_representation} and
rearranging the summations gives
\begin{equation}
\bar{\mu}(\mathbf{x})=\sum_{i=1}^{K}\left[\sum_{m=1}^{M}w_m\alpha_i(s_m)\right]G_i(\mathbf{x})=\sum_{i=1}^{K}\bar{\alpha}_iG_i(\mathbf{x}),
\label{eq:spectral_aggregation_closure}
\end{equation}
where
$\bar{\alpha}_i=\sum_{m=1}^{M}w_m\alpha_i(s_m)$.
Thus, linear spectral aggregation changes only the Gaussian density
coefficients and preserves the shared geometry. Overall, the
shared geometry representation confines spectral variation to
Gaussian density changes and preserves the same spatial primitives
under both spectral differencing and linear aggregation.

\subsection{Shared-Structure 4D-SG Reconstruction}
\label{sec:framework}

An overview of the framework is shown in Fig.~\ref{fig:framework}. Let $\Theta=\{\theta_n\}_{n=1}^{N_\theta}$ denote the sparse projection views and $\mathcal{S}=\{s_m\}_{m=1}^{M}$ denote the observed spectral channels. The input projections are denoted by $\mathcal{P}=\{P_{\theta}^{(s_m)}\mid\theta\in\Theta,s_m\in\mathcal{S}\}$. As illustrated in Fig.~\ref{fig:framework}, the projections at each view are aggregated into a structural projection over the full spectrum for establishing a common Gaussian representation. The learned Gaussian geometry is reused across spectral channels, while GSC-Net models the raw density curves for each Gaussian used for reconstruction of observed spectral channels and prediction of unobserved spectral channels.

The common Gaussian representation is defined as
\begin{equation}
\mathcal{G}^{(c)}
=
\left\{
\left(
\boldsymbol{\mu}_i,
\mathbf{s}_i,
\mathbf{q}_i,
u_i^{(c)}
\right)
\right\}_{i=1}^{K},
\label{eq:canonical_gaussians}
\end{equation}
where $\boldsymbol{\mu}_i$, $\mathbf{s}_i$, $\mathbf{q}_i$, and $u_i^{(c)}$ denote the Gaussian center, scale, rotation quaternion, and common raw density, respectively. The centers, scales, and rotations constitute the shared Gaussian geometry, and the corresponding common density coefficient is $\alpha_i^{(c)}=\sigma_+(u_i^{(c)})$.

For each projection view $\theta$, the structural projection over the full spectrum is obtained as $\bar{P}_{\theta}=\mathcal{A}_{\mathrm{fs}}(\{P_{\theta}^{(s_m)}\}_{s_m\in\mathcal{S}})$. For photon-counting CT, $\mathcal{A}_{\mathrm{fs}}(\cdot)$ returns the projection over the full spectrum, whereas for conventional spectral CT, it computes the average projection as $\bar{P}_{\theta}=\frac{1}{M}\sum_{m=1}^{M}P_{\theta}^{(s_m)}$. Both forms are used to emphasize the structural information shared across spectral channels. The common Gaussian representation is then supervised through
\begin{equation}
\resizebox{0.98\columnwidth}{!}{$\displaystyle
\min_{\mathcal{G}^{(c)}}\mathcal{L}_{\mathrm{geom}}
=
\frac{1}{|\Theta|}
\sum_{\theta\in\Theta}
\mathcal{L}_{\mathrm{img}}
\left(
\mathcal{R}_{\theta}(\mathcal{G}^{(c)}),
\bar{P}_{\theta}
\right)
+
\lambda_{\mathrm{tv}}
\mathcal{L}_{\mathrm{tv}}^{(c)}
$}
\label{eq:canonical_geometry_loss}
\end{equation}
Here, $\mathcal{R}_{\theta}(\cdot)$ and $\mathcal{Q}(\cdot)$ denote the Gaussian rendering and voxelization operators, respectively, following R$^2$-Gaussian \cite{zha2024r2gaussian}, and $\mathcal{L}_{\mathrm{img}}$ consists of an $\ell_1$ term and an optional SSIM term. Adaptive densification and pruning adjust the allocation of Gaussian primitives during this optimization.

Once the common Gaussian representation is established, the centers, scales, and rotations $\{(\boldsymbol{\mu}_i,\mathbf{s}_i,\mathbf{q}_i)\}_{i=1}^{K}$ are fixed to preserve Gaussian correspondence across spectral channels. The common raw densities and GSC-Net remain learnable. For the observed spectral grid, GSC-Net predicts the raw density transformation vector of each Gaussian:
\begin{equation}
\Delta\mathbf{u}_i
=
\left[
\Delta u_i^{(1)},\ldots,\Delta u_i^{(M)}
\right].
\label{eq:framework_spectral_transform}
\end{equation}
For spectral channel $s_m$, its raw density is obtained as
\begin{equation}
u_i^{(m)}
=
u_i^{(c)}
+
\Delta u_i^{(m)}.
\label{eq:framework_raw_density}
\end{equation}
The corresponding non-negative density coefficient is
\begin{equation}
\alpha_i^{(m)}
=
\sigma_+\left(u_i^{(m)}\right).
\label{eq:framework_density_mapping}
\end{equation}
The entries of $\Delta\mathbf{u}_i$ form the raw density curve of each Gaussian over the observed spectral grid. For spectral channel $s_m$, the corresponding Gaussian representation is $\hat{\mathcal{G}}^{(s_m)}=\{(\boldsymbol{\mu}_i,\mathbf{s}_i,\mathbf{q}_i,\alpha_i^{(m)})\}_{i=1}^{K}$. Its rendered projection and queried attenuation volume are denoted by $\hat{P}_{\theta}^{(s_m)}=\mathcal{R}_{\theta}(\hat{\mathcal{G}}^{(s_m)})$ and $\hat{V}^{(s_m)}=\mathcal{Q}(\hat{\mathcal{G}}^{(s_m)})$, respectively.

At each iteration, a projection view is sampled and all available spectral channels at that view contribute to the projection supervision. The spectral optimization objective is
\begin{equation}
\begin{split}
\mathcal{L}_{\mathrm{spec}}
=
&\mathbb{E}_{\theta\sim\Theta}
\left[
\frac{1}{|\mathcal{S}_{\theta}|}
\sum_{s_m\in\mathcal{S}_{\theta}}
\mathcal{L}_{\mathrm{img}}
\left(
\hat{P}_{\theta}^{(s_m)},
P_{\theta}^{(s_m)}
\right)
\right] \\
&+
\lambda_{\mathrm{curve}}
\mathcal{L}_{\mathrm{curve}}
+
\lambda_{\mathrm{tv}}
\mathcal{L}_{\mathrm{tv}}.
\end{split}
\label{eq:spectral_curve_loss}
\end{equation}
Here, $\mathcal{S}_{\theta}$ denotes the spectral channels available at view $\theta$. The term $\mathcal{L}_{\mathrm{curve}}$ regularizes the raw density curves of individual Gaussians along the ordered spectral axis, while $\mathcal{L}_{\mathrm{tv}}$ regularizes the queried attenuation volumes. For an observed spectral channel, the corresponding raw density transformation is queried directly from the learned curve. For an unobserved target spectral coordinate $s^\star$ with normalized coordinate $\tau^\star$, $\Delta u_i(\tau^\star)$ is queried from the learned raw density curve and combined with the same shared Gaussian geometry according to Eq.~\eqref{eq:continuous_4d_sg_representation}. Algorithm~\ref{alg:framework} summarizes the optimization and spectral querying procedure.

\begin{algorithm}[t]
\caption{Optimization and spectral querying of 4D-SG}
\label{alg:framework}
\begin{algorithmic}[1]
\Require $\mathcal{P}$, $\Theta$, observed grid $\mathcal{S}$, target $(s^\star,\tau^\star)$
\Ensure $\{\hat{V}^{(s_m)}\}_{m=1}^{M}$ and $\hat{V}^{(s^\star)}$

\State Construct $\{\bar{P}_{\theta}\}_{\theta\in\Theta}$ and initialize
$\mathcal{G}^{(c)}$ as described in Sec.~\ref{sec:framework}

\Repeat
    \State Optimize $\mathcal{G}^{(c)}$ using
    Eq.~\eqref{eq:canonical_geometry_loss}, including scheduled densification and pruning
\Until{convergence}

\State Fix the shared Gaussian geometry

\Repeat
    \State Sample $\theta\sim\Theta$ and predict the spectral curves using GSC-Net
    \State Render $\{s_m\in\mathcal{S}_{\theta}\}$ and optimize the common raw
    densities and GSC-Net using Eq.~\eqref{eq:spectral_curve_loss}
\Until{convergence}

\State Query $\{\hat{V}^{(s_m)}\}_{m=1}^{M}$ using
Eqs.~\eqref{eq:framework_raw_density}--\eqref{eq:framework_density_mapping}

\State Query $\hat{V}^{(s^\star)}$ using
Eq.~\eqref{eq:target_transform_interpolation}

\end{algorithmic}
\end{algorithm}

\begin{table*}[!t]
    \centering
    \caption{Summary of the datasets, spectral configurations, and acquisition geometries. Chest, Head, Walnut, and Ordinary Objects use the same simulated cone-beam system matrix, while full-view FDK reconstructions are used as reference volumes for Insect and Mouse Chest.}
    \label{tab:dataset_summary}
    \fontsize{6.7pt}{8.1pt}\selectfont
    \setlength{\tabcolsep}{3.2pt}
    \renewcommand{\arraystretch}{1.10}
    \begin{tabularx}{\textwidth}{@{}l >{\raggedright\arraybackslash}X >{\raggedright\arraybackslash}X l >{\raggedright\arraybackslash}p{2.6cm} >{\raggedright\arraybackslash}p{2.1cm}@{}}
        \toprule
        \textbf{Dataset}
        & \textbf{Data Source}
        & \textbf{Spectral Setting}
        & \textbf{\shortstack[l]{SOD/SDD(mm)}}
        & \textbf{Detector}
        & \textbf{Reference Volume} \\
        \midrule
        Chest~\cite{aapm2017lowdose} & Synthesized from a single-energy CT volume & 3 monoenergetic channels: 40, 70, and 100~keV & 1000/1500 & $512\times512$ pixels, 1-mm pixel size & Synthesized GT \\
        Head~\cite{klacansky2022scientific} & Synthesized from a single-energy CT volume & 3 monoenergetic channels: 40, 70, and 100~keV & 1000/1500 & $512\times512$ pixels, 1-mm pixel size & Synthesized GT \\
        Walnut~\cite{zhou2025cone} & Simulated from a spectral CT volume & 8 monoenergetic channels: 10--80~keV at 10-keV intervals & 1000/1500 & $512\times512$ pixels, 1-mm pixel size & Spectral CT volume \\
        Ordinary Objects~\cite{volgyes2018dual} & Simulated from dual-energy CT volumes & 5 mixed-energy channels from 100/140-kVp volumes, with $\lambda=\{0,0.3,0.5,0.7,1.0\}$ & 1000/1500 & $512\times512$ pixels, 1-mm pixel size & Spectral CT volume \\
        Insect & Real multi-kVp cone-beam CT projections & 4 tube-voltage settings: 40, 60, 80, and 100~kVp & 310/510 & $2882\times2340$ pixels, 0.05-mm pixel size & Full-view FDK \\
        Mouse Chest & Real photon-counting CT projections & 4 energy windows: 20--90, 30--90, 45--90, and 60--90~keV & 76.28/361.1 & $2062\times2062$ pixels, 0.1-mm pixel size & Full-view FDK \\
        \bottomrule
    \end{tabularx}
\end{table*}

\begin{table*}[t]
    \centering
    \caption{Quantitative comparison on all datasets using 50 views. 
    Higher PSNR and SSIM values are better, whereas lower LPIPS values are better.
    Metrics on Insect and Mouse Chest are computed with respect to full-view FDK references.}
    \label{tab:quant_all_50view}
    \scriptsize
    \setlength{\tabcolsep}{1.8pt}
    \renewcommand{\arraystretch}{1.12}
    \resizebox{\textwidth}{!}{
    \begin{tabular}{lccccccccccccccccccccc}
        \toprule
        \multicolumn{1}{c}{\textbf{Method}}
        & \multicolumn{3}{c}{\textbf{Chest}}
        & \multicolumn{3}{c}{\textbf{Head}}
        & \multicolumn{3}{c}{\textbf{Walnut}}
        & \multicolumn{3}{c}{\textbf{Ordinary Objects}}
        & \multicolumn{3}{c}{\textbf{Insect}}
        & \multicolumn{3}{c}{\textbf{Mouse Chest}}
        & \multicolumn{3}{c}{\textbf{Average}} \\
        \cmidrule(lr){2-4}
        \cmidrule(lr){5-7}
        \cmidrule(lr){8-10}
        \cmidrule(lr){11-13}
        \cmidrule(lr){14-16}
        \cmidrule(lr){17-19}
        \cmidrule(lr){20-22}
        & PSNR$\uparrow$ & SSIM$\uparrow$ & LPIPS$\downarrow$
        & PSNR$\uparrow$ & SSIM$\uparrow$ & LPIPS$\downarrow$
        & PSNR$\uparrow$ & SSIM$\uparrow$ & LPIPS$\downarrow$
        & PSNR$\uparrow$ & SSIM$\uparrow$ & LPIPS$\downarrow$
        & PSNR$\uparrow$ & SSIM$\uparrow$ & LPIPS$\downarrow$
        & PSNR$\uparrow$ & SSIM$\uparrow$ & LPIPS$\downarrow$
        & PSNR$\uparrow$ & SSIM$\uparrow$ & LPIPS$\downarrow$ \\
        \midrule

        FDK
        & 27.00 & 0.546 & 0.402
        & 29.41 & 0.702 & 0.346
        & 24.77 & 0.540 & 0.393
        & 29.85 & 0.625 & 0.326
        & 27.09 & 0.526 & 0.440
        & 22.61 & 0.516 & 0.432
        & 26.79 & 0.576 & 0.390 \\

        CGLS
        & 33.43 & 0.858 & 0.244
        & 35.08 & 0.909 & 0.256
        & 29.99 & 0.813 & 0.283
        & 37.74 & 0.829 & 0.243
        & 30.50 & 0.877 & 0.239
        & 27.43 & 0.669 & 0.419
        & 32.36 & 0.826 & 0.281 \\

        SART
        & 32.85 & 0.886 & 0.228
        & 36.73 & 0.925 & 0.256
        & 32.76 & 0.897 & 0.244
        & 39.26 & 0.943 & 0.225
        & 31.96 & 0.900 & 0.236
        & 28.78 & 0.705 & 0.361
        & 33.72 & 0.876 & 0.258 \\

        NAF
        & 35.56 & 0.896 & 0.192
        & 37.35 & 0.942 & 0.215
        & 33.68 & 0.939 & 0.199
        & 40.12 & 0.968 & 0.193
        & 32.74 & 0.914 & 0.194
        & 28.80 & 0.715 & 0.349
        & 34.71 & 0.896 & 0.224 \\

        R$^2$-Gaussian
        & 35.77 & 0.911 & 0.174
        & 38.15 & 0.947 & 0.209
        & 34.93 & 0.948 & 0.186
        & 40.34 & \textbf{0.974} & \textbf{0.170}
        & 35.06 & 0.924 & 0.146
        & 29.08 & \textbf{0.751} & 0.360
        & 35.56 & 0.909 & 0.208 \\

        4D-SG (Ours)
        & \textbf{37.57} & \textbf{0.923} & \textbf{0.162}
        & \textbf{39.54} & \textbf{0.959} & \textbf{0.203}
        & \textbf{35.65} & \textbf{0.955} & \textbf{0.177}
        & \textbf{41.04} & \textbf{0.974} & 0.179
        & \textbf{36.54} & \textbf{0.936} & \textbf{0.112}
        & \textbf{29.35} & 0.736 & \textbf{0.331}
        & \textbf{36.61} & \textbf{0.914} & \textbf{0.194} \\

        \bottomrule
    \end{tabular}}
\end{table*}

\subsection{Gaussian-wise Spectral Density Curve Network}
\label{sec:gscnet}

As illustrated in Fig.~\ref{fig:net}, GSC-Net maps the common raw density $u_i^{(c)}$ and the normalized spectral grid values $\{\tau_m\}_{m=1}^{M}$ to the raw density transformations $\{\Delta u_i^{(m)}\}_{m=1}^{M}$ for each Gaussian. Here, $\tau_m$ denotes the normalized spectral coordinate associated with $s_m$. The predicted transformations form a raw density curve over the ordered spectral axis, while the shared Gaussian geometry remains fixed.

\textbf{Tokenization: }
The common raw density and each normalized spectral grid value are encoded separately and combined by broadcasted addition:
\begin{equation}
\mathbf{x}_{i,m}^{(0)}
=
\gamma_u\left(u_i^{(c)}\right)
+
\gamma_s(\tau_m).
\label{eq:gscnet_token}
\end{equation}
The resulting tokens are arranged according to the ordered spectral grid:
\begin{equation}
\mathbf{X}_i^{(0)}
=
\left[
\mathbf{x}_{i,1}^{(0)},\ldots,\mathbf{x}_{i,M}^{(0)}
\right]
\in\mathbb{R}^{M\times d}.
\label{eq:gscnet_tokenization}
\end{equation}
The encoders $\gamma_u(\cdot)$ and $\gamma_s(\cdot)$ include the corresponding embedding, linear projection, and activation operations shown in Fig.~\ref{fig:net}. The resulting sequence preserves the order of the spectral grid values and is conditioned on the common raw density of each Gaussian.

\textbf{Spectral SSM Backbone: }
The token sequence is processed along the spectral axis by a spectral state space model (SSM) backbone:
\begin{equation}
\begin{aligned}
\mathbf{H}_i
&=
\left[
\mathbf{h}_{i,1},\ldots,\mathbf{h}_{i,M}
\right] \\
&=
\Phi_{\mathrm{post}}
\left(
\Phi_{\mathrm{ssm}}
\left(
\Phi_{\mathrm{conv}}
\left(
\mathbf{X}_i^{(0)}
\right)
\right)
\odot
\Phi_{\mathrm{gate}}
\left(
\mathbf{X}_i^{(0)}
\right)
\right).
\end{aligned}
\label{eq:gscnet_backbone}
\end{equation}
Here, $\Phi_{\mathrm{conv}}(\cdot)$ captures local interactions between neighboring spectral tokens, $\Phi_{\mathrm{ssm}}(\cdot)$ models their ordered dependencies, and $\Phi_{\mathrm{gate}}(\cdot)$ modulates the resulting features. The output projection $\Phi_{\mathrm{post}}(\cdot)$ produces the spectral features of the tokens. A global feature $\mathbf{p}_i=\operatorname{Pool}(\mathbf{H}_i)$ is further obtained for curve decoding.

\textbf{Density Curve Head: }
The density curve head combines token, low-rank basis, and residual responses. The token response $\delta_{i,m}^{\mathrm{tok}}$ is decoded from $\mathbf{h}_{i,m}$, while the pooled feature $\mathbf{p}_i$ provides the coefficients of the basis response and the residual response. The low-rank basis response is
\begin{equation}
\delta_{i,m}^{\mathrm{basis}}
=
\sum_{r=1}^{R}
a_{i,r}B_{r,m}.
\label{eq:gscnet_basis_response}
\end{equation}
The raw density transformation for each Gaussian is obtained as
\begin{equation}
\Delta u_i^{(m)}
=
\delta_{i,m}^{\mathrm{tok}}
+
\delta_{i,m}^{\mathrm{basis}}
+
\delta_{i,m}^{\mathrm{res}}.
\label{eq:gscnet_curve_head}
\end{equation}
Here, $\mathbf{B}\in\mathbb{R}^{R\times M}$ is a learnable spectral basis matrix, $R$ denotes the basis rank, and $a_{i,r}$ denotes the corresponding coefficient for each Gaussian. For each channel, the raw density and non-negative density coefficient are obtained as $u_i^{(m)}=u_i^{(c)}+\Delta u_i^{(m)}$ and $\alpha_i^{(m)}=\sigma_+(u_i^{(m)})$, respectively.

For an observed spectral channel $s_m$, the corresponding transformation $\Delta u_i^{(m)}$ is queried directly from the predicted curve. For an unobserved target spectral channel, the raw density transformation is queried from the predicted curve as described in Sec.~\ref{sec:spectral_interpolation}, and the resulting raw density is mapped to the corresponding density coefficient through the softplus activation.

A first-order smoothness term constrains adjacent values of the raw density curve of each Gaussian:
\begin{equation}
\mathcal{L}_{\mathrm{curve}}
=
\frac{1}{K(M-1)}
\sum_{i=1}^{K}
\sum_{m=1}^{M-1}
\left|
\Delta u_i^{(m+1)}
-
\Delta u_i^{(m)}
\right|.
\label{eq:curve_regularizer}
\end{equation}
This term suppresses abrupt variation between neighboring spectral grid values while preserving the spectral responses of individual Gaussians.

\begin{figure*}[!t]
    \centering
    \includegraphics[width=\textwidth]{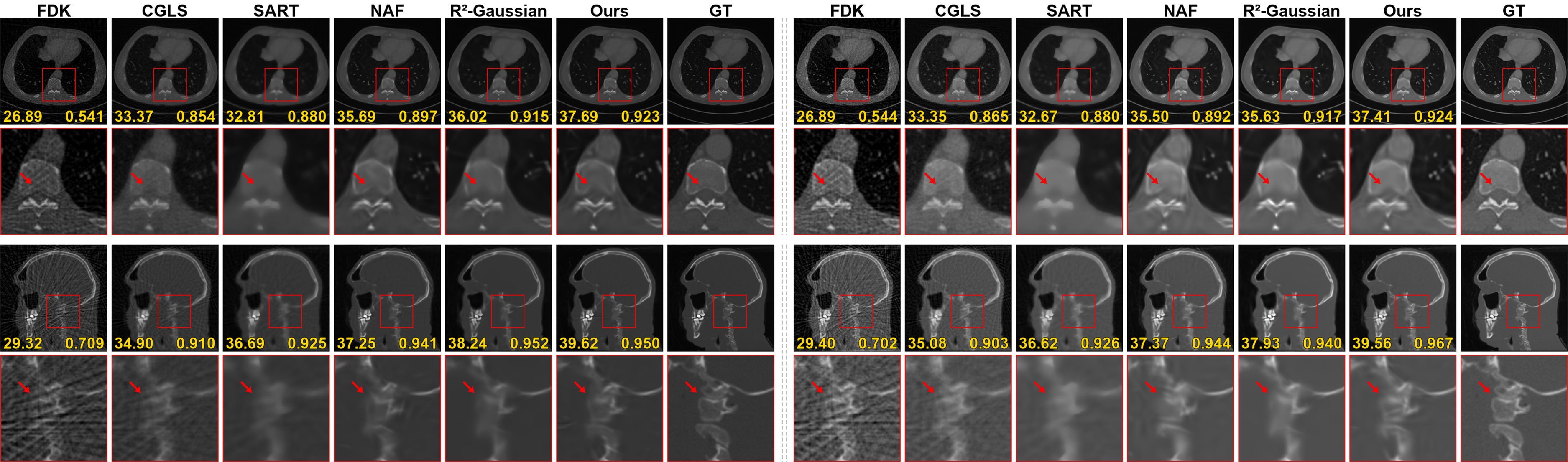}
    \caption{Representative qualitative comparison on 3D volume datasets with simulated projections using 50 views. 
    The upper two groups show Walnut at 30~keV and 70~keV, respectively, and the lower two groups show Ordinary Objects with mixing coefficients of 0.0 and 1.0, respectively. 
    For each spectral setting, the first row shows the reconstructed slice and the second row shows the corresponding zoomed ROI. 
    The red boxes indicate the enlarged regions, and the red arrows highlight representative local structures. 
    The PSNR and SSIM values annotated in each panel are computed on the corresponding entire 3D volume at the displayed spectral channel, rather than on the shown slice only. HU values are normalized to [0,1].}
    \label{fig:qual_sim_50view}
\end{figure*}
\begin{figure*}[!t]
    \centering
    \includegraphics[width=\textwidth]{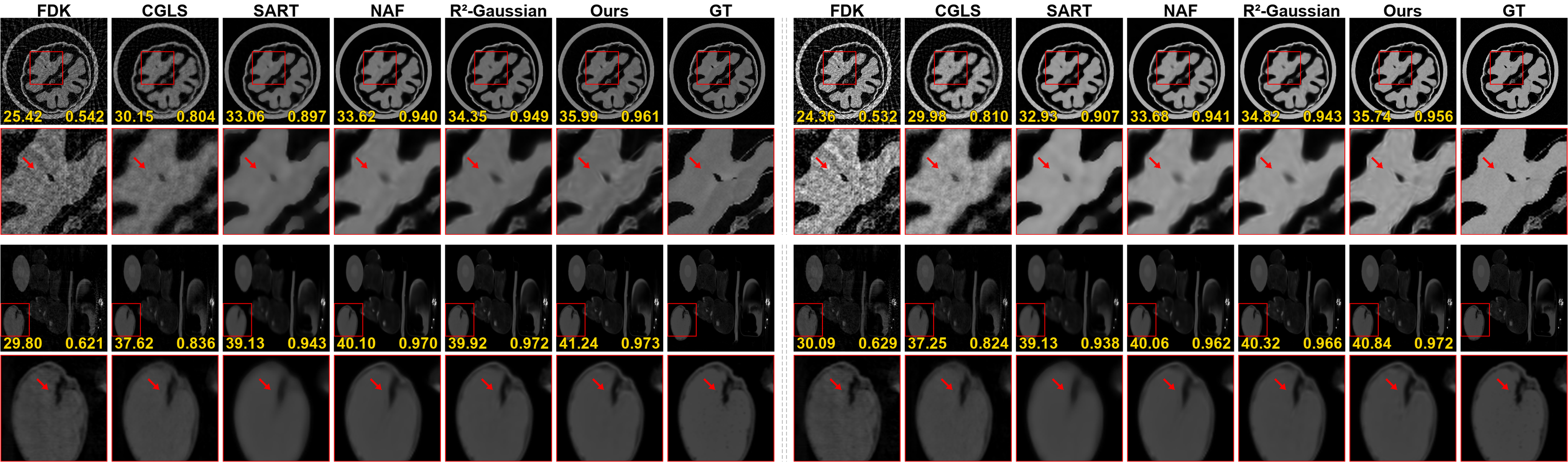}
    \caption{Representative qualitative comparison on synthesized multi-energy datasets using 50 views. 
    The upper two groups show Chest at 40~keV and 100~keV, respectively, and the lower two groups show Head at 40~keV and 100~keV, respectively. 
    For each spectral setting, the first row shows the reconstructed slice and the second row shows the corresponding zoomed ROI. 
    The red boxes indicate the enlarged regions, and the red arrows highlight representative local structures. 
    The PSNR and SSIM values annotated in each panel are computed on the corresponding entire 3D volume at the displayed spectral channel, rather than on the shown slice only. HU values are normalized to [0,1].}
    \label{fig:qual_synth_50view}
\end{figure*}

\subsection{Querying Unobserved Spectral Channels}
\label{sec:spectral_interpolation}
Given an unobserved target spectral channel $s^\star$, let $\tau^\star$ denote its normalized spectral grid value. For $\tau_k\leq\tau^\star\leq\tau_{k+1}$, the raw density transformation of the $i$-th Gaussian is queried from the learned curve using piecewise-linear interpolation:
\begin{equation}
\Delta u_i(\tau^\star)
=
\frac{\tau_{k+1}-\tau^\star}{\tau_{k+1}-\tau_k}
\Delta u_i^{(k)}
+
\frac{\tau^\star-\tau_k}{\tau_{k+1}-\tau_k}
\Delta u_i^{(k+1)}.
\label{eq:target_transform_interpolation}
\end{equation}

The queried transformation is combined with the common raw density and mapped to a non-negative density coefficient using the raw density parameterization in Eqs.~\eqref{eq:framework_raw_density}--\eqref{eq:framework_density_mapping}. The resulting coefficient is combined with the shared Gaussian geometry to query the target attenuation volume using the same Gaussian voxelization procedure described in Sec.~\ref{sec:framework}. No additional optimization is required.

\section{Experiments}
\subsection{Datasets and Experimental Settings}

The proposed method is evaluated on six datasets spanning synthesized data, data with simulated projections, and real projection measurements, as summarized in Table~\ref{tab:dataset_summary}. Unless otherwise specified, all experiments are conducted using 50 projection views. For the datasets with simulated projections, the views are uniformly distributed over $360^\circ$. For the real projection datasets, 50 views are uniformly selected from the original full-view measurements. All reconstructed volumes are resized or cropped to a spatial resolution of $256^3$ for evaluation. The spectral channels of the Insect dataset are labeled in kVp because each tube voltage setting corresponds to a polychromatic rather than monoenergetic X-ray spectrum.

The proposed method is implemented in PyTorch, and all experiments are performed on an NVIDIA GeForce RTX 5070 Ti GPU. The source code is available at \href{https://github.com/yqx7150/4D-SG}{https://github.com/yqx7150/4D-SG}. The optimization consists of two stages. We first optimize the shared Gaussian geometry for 30{,}000 iterations and then optimize the raw density curves for each Gaussian for 5{,}000 iterations. The Gaussian primitives are initialized from an FDK reconstruction, and the same hyperparameter settings are used for all datasets unless a dataset is specified otherwise.

\begin{figure*}[!t]
    \centering
    \includegraphics[width=\textwidth]{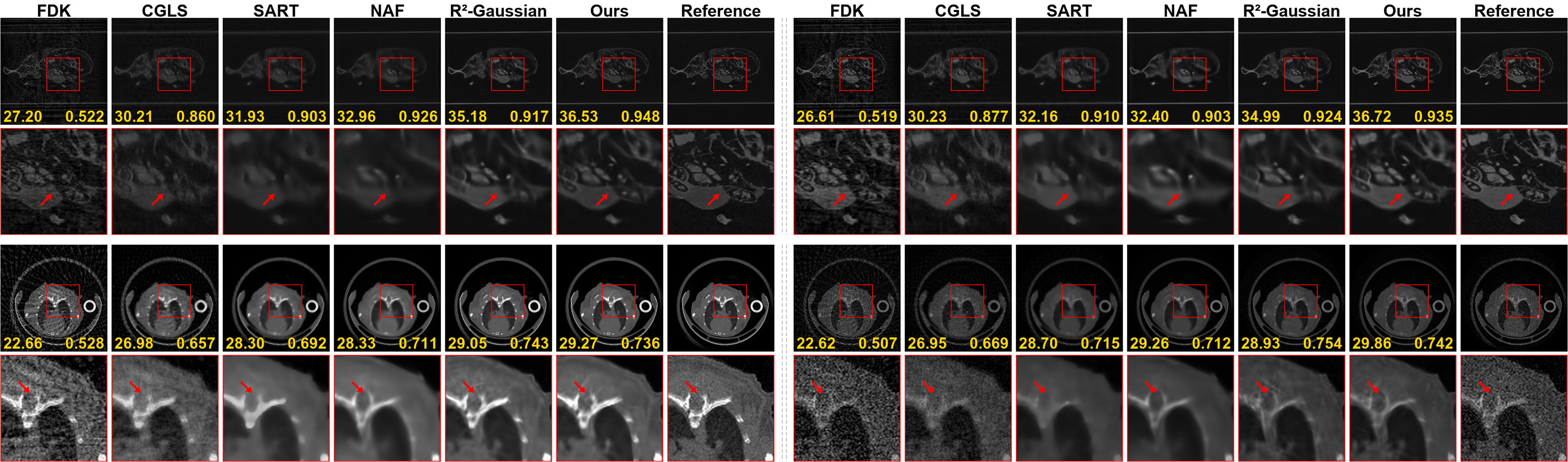}
    \caption{Representative qualitative comparison on real projection datasets using 50 views. 
    The upper two groups show Insect acquired at 40~kVp and 100~kVp, respectively, and the lower two groups show Mouse Chest at the 20--90~keV and 60--90~keV energy windows, respectively. 
    For each spectral setting, the first row shows the reconstructed slice and the second row shows the corresponding zoomed ROI. 
    The red boxes indicate the enlarged regions, and the red arrows highlight representative local structures. 
    Since clean ground-truth volumes are unavailable for the real projection datasets, the reference volumes are full-view FDK reconstructions. 
    The PSNR and SSIM values annotated in each panel are computed on the corresponding entire 3D volume at the displayed spectral channel, rather than on the shown slice only.}
    \label{fig:qual_real_50view}
\end{figure*}

\subsection{Comparative Results}

\begin{figure}[!t]
    \centering
    \includegraphics[width=\columnwidth]{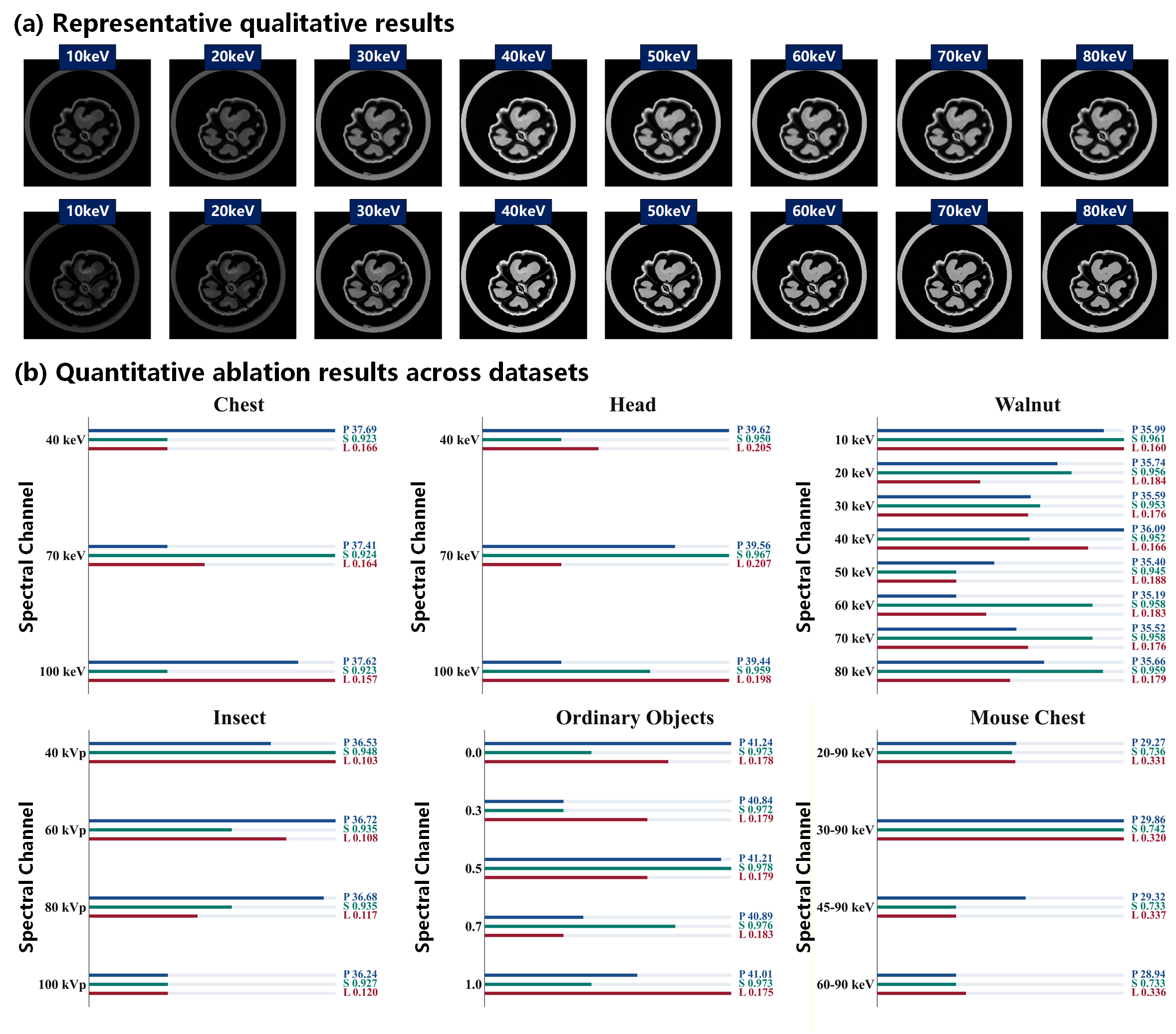}
    \caption{Reconstruction and quantitative evaluation across spectral channels. (a) Reconstructed and reference Walnut slices. (b) PSNR, SSIM, and LPIPS by channel for each dataset.}
    \label{fig:energy_slice_metrics}
\end{figure}

The proposed method is compared with FDK, CGLS, SART, NAF~\cite{zha2022naf}, and R$^2$-Gaussian~\cite{zha2024r2gaussian}. Since NAF and R$^2$-Gaussian do not use spectral reconstruction with shared structure, they are applied to each spectral channel independently. PSNR, SSIM, and LPIPS are averaged over all reconstructed slices and spectral channels after intensity normalization. For the qualitative figures, the annotated PSNR and SSIM values are computed on the corresponding entire 3D volume at the displayed spectral setting.

Across Figs.~\ref{fig:qual_sim_50view}--\ref{fig:qual_real_50view}, FDK exhibits severe sparse-view artifacts, while CGLS and SART reduce global artifacts but blur or distort fine structures. NAF and R$^2$-Gaussian further improve structural continuity, whereas 4D-SG preserves sharper boundaries, local contrast, and structural consistency across channels on synthesized data and data with simulated projections. The same advantage remains visible on real projection data despite acquisition noise, system blur, and mismatch in spectral response.

Table~\ref{tab:quant_all_50view} reports the results using 50 views. The proposed method achieves the best average PSNR, SSIM, and LPIPS. On synthesized and simulated projection datasets, the improvements indicate better fidelity to the available reference volumes. On Insect and Mouse Chest, metrics use full-view FDK references and should be interpreted as reference comparisons. Overall, the results consistently indicate improved sparse-view spectral CT reconstruction across synthesized, simulated projection, and real projection scenarios.

\begin{figure}[!t]
    \centering
    \includegraphics[width=\columnwidth]{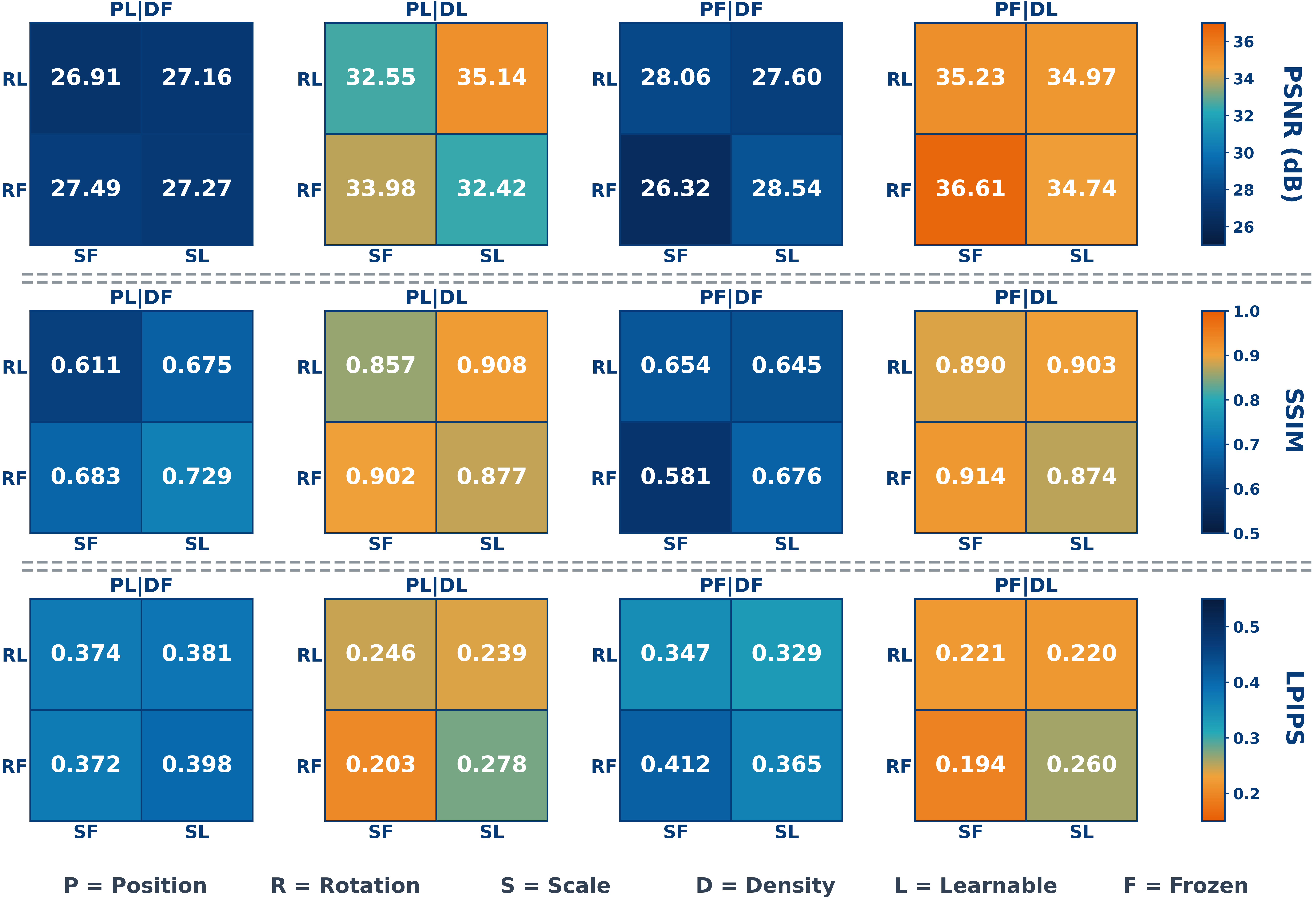}
    \caption{Ablation on Gaussian parameter freezing during raw density curve learning. Rows report PSNR, SSIM, and LPIPS. P/R/S/D denote position, rotation, scale, and raw density; L/F denote learnable/frozen.}
    \label{fig:freeze_strategy}
\end{figure}

\subsection{Ablation Studies}
Channel reconstruction, Gaussian parameter freezing strategies, and the contribution of GSC-Net are examined. Scalability with respect to spectral channel count and interpolation at unobserved channels are also evaluated.

\textbf{Spectral Channel Reconstruction and Metric Analysis by Dataset: }
Fig.~\ref{fig:energy_slice_metrics} shows that the reconstructed slices preserve object boundaries and internal structures across spectral channels while reproducing spectral contrast changes. Stable PSNR, SSIM, and LPIPS across data types further indicate that the learned Gaussian raw density curves coherently represent attenuation variation along the spectral axis.

\textbf{Effect of Freezing Different Gaussian Parameters: }
After learning the shared Gaussian geometry, we evaluate several parameter freezing strategies during the subsequent spectral density curve learning stage, as shown in Fig.~\ref{fig:freeze_strategy}.

Freezing the raw density leads to a substantial degradation in reconstruction quality. The best performance is obtained when the Gaussian positions, rotations, and scales are fixed, while the raw density remains learnable, achieving $36.61$ dB PSNR, $0.914$ SSIM, and $0.194$ LPIPS. Allowing the geometric parameters to remain learnable does not provide any further improvement, which supports the use of fixed shared Gaussian geometry during spectral density curve learning.

\begin{figure}[!t]
    \centering
    \includegraphics[width=\columnwidth]{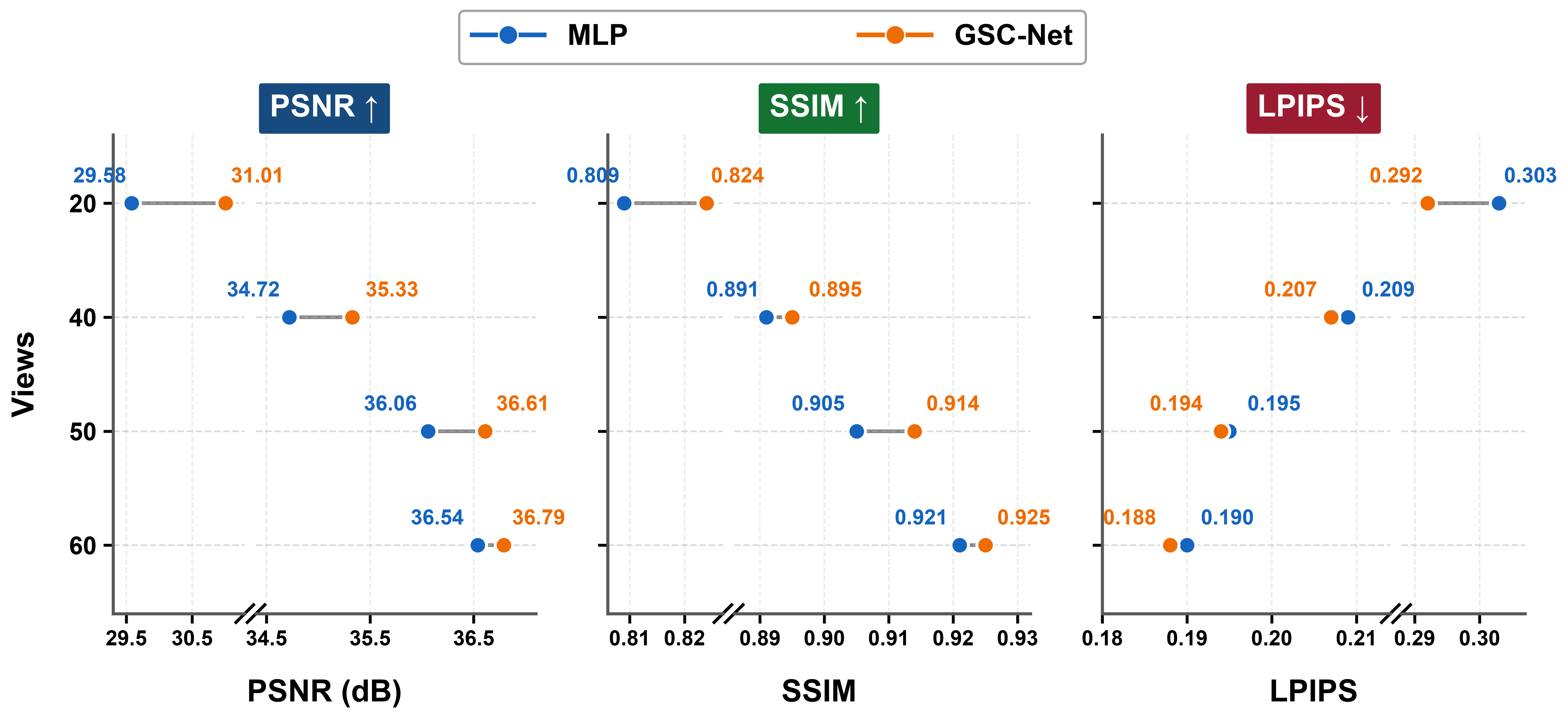}
    \caption{Comparison between GSC-Net and a direct MLP alternative under different numbers of projection views. Higher values are better for PSNR and SSIM, whereas lower values are better for LPIPS.}
    \label{fig:gscnet_mlp_view_ablation}
\end{figure}

\begin{figure}[!t]
    \centering
    \includegraphics[width=\columnwidth]{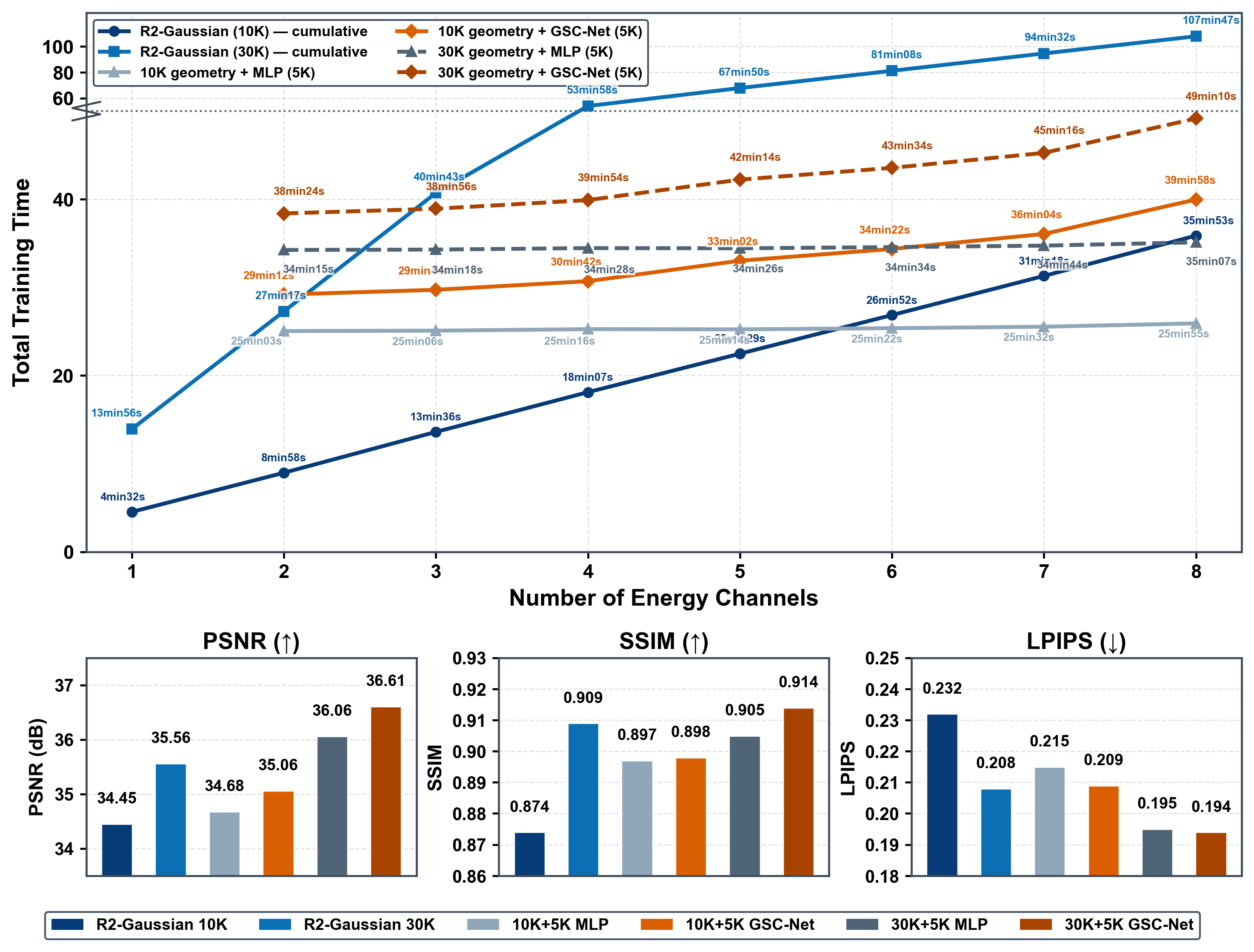}
    \caption{Comparison of training time and reconstruction quality for different spectral channel counts and configurations. The top plot reports how total training time changes as the number of spectral channels increases. It compares independent R$^2$-Gaussian training against shared geometry variants using either MLP or GSC-Net for spectral adaptation. The bottom plots report PSNR, SSIM, and LPIPS. These results indicate that reusing shared Gaussian geometry improves training efficiency while maintaining reconstruction accuracy.}
    \label{fig:time_vs_channels}
\end{figure}

\begin{figure}[!t]
    \centering
    \includegraphics[width=\columnwidth]{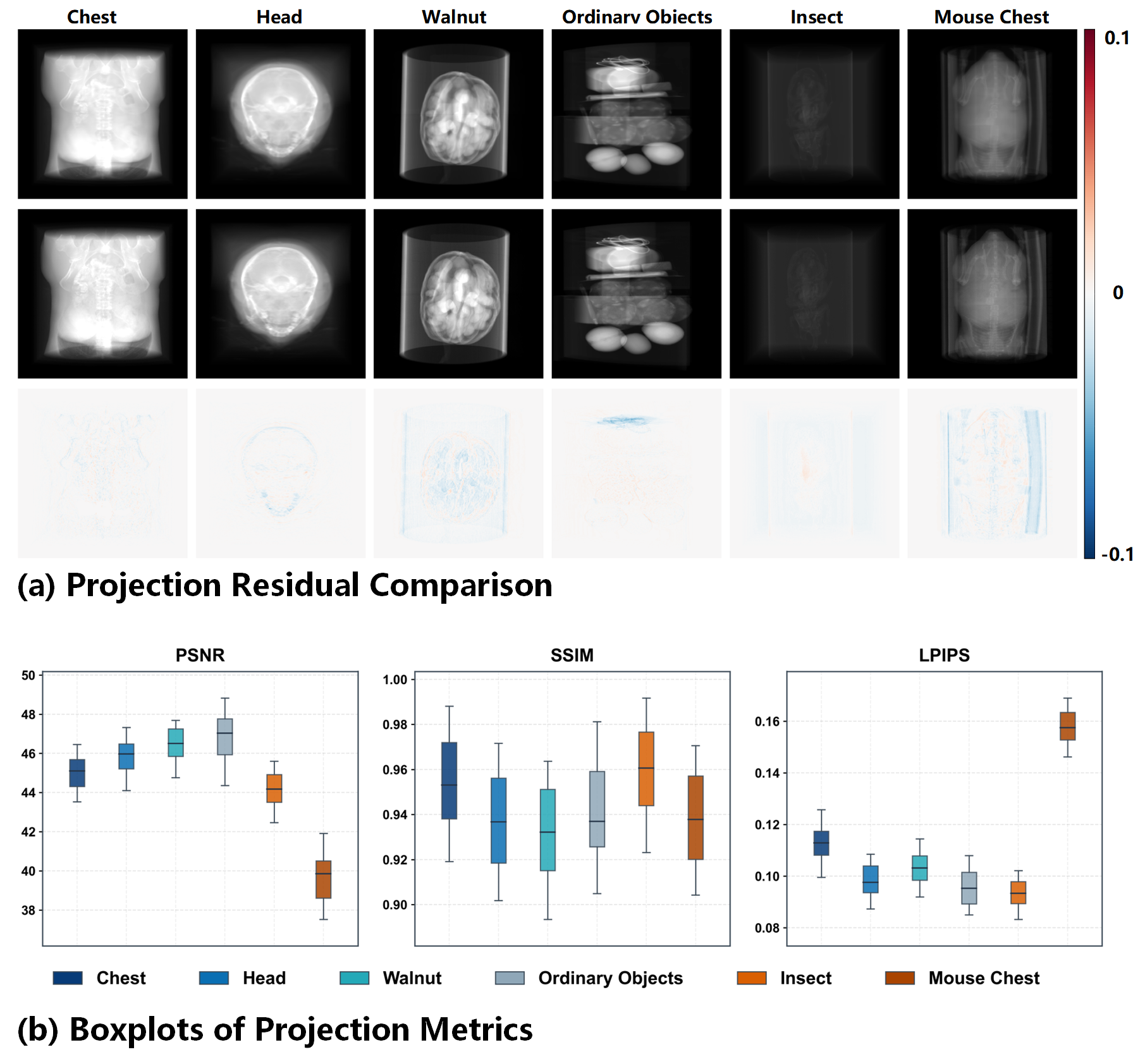}
    \caption{Prediction of projections for an unobserved spectral channel across six datasets. (a) Predicted projections, corresponding reference projections, and residual maps. (b) Projection-level metrics evaluated across multiple views.}
    \label{fig:novel_spectrum}
\end{figure}

\textbf{Effect of GSC-Net and Number of Views: }
We evaluate the effect of GSC-Net by replacing it with a direct MLP while keeping the remaining pipeline unchanged. We also compare the two variants under different numbers of projection views. The results are shown in Fig.~\ref{fig:gscnet_mlp_view_ablation}.

GSC-Net outperforms the direct MLP at all tested view counts. At 50 views, it improves PSNR from $36.06$ to $36.61$ dB and SSIM from $0.905$ to $0.914$, while reducing LPIPS from $0.195$ to $0.194$. At 20 views, PSNR increases from $29.58$ to $31.01$ dB and LPIPS decreases from $0.303$ to $0.292$. Thus, modeling Gaussian raw density curves is more effective than direct density regression, particularly when angular sampling is severely limited.

\textbf{Training Efficiency versus Number of Spectral Channels: }
We compare independent channel training with R$^2$-Gaussian against MLP and GSC-Net variants with shared Gaussian geometry in training time and quality (Fig.~\ref{fig:time_vs_channels}).

Independent R$^2$-Gaussian training grows rapidly with spectral channel count, whereas shared geometry variants scale more slowly. For eight channels, R$^2$-Gaussian with 30K iterations requires 107~min~47~s, compared with 49~min~10~s for 30K geometry + GSC-Net. This variant gives the best shared geometry result ($36.61$ dB PSNR, $0.914$ SSIM, and $0.194$ LPIPS), reducing redundant channel optimization without sacrificing accuracy.

\textbf{Unobserved Spectral Channel Projection Prediction: }
\label{sec:unobserved_energy_projection}
We evaluate target spectral channel prediction by withholding one intermediate spectral channel during spectral density curve learning. The remaining channels are used as observed channels, and the withheld projection is generated by querying the learned raw density transformation curves for each Gaussian.

Across the six datasets shown in Fig.~\ref{fig:novel_spectrum}, predicted projections closely match the ground-truth or reference projections, with residuals mainly near high-contrast boundaries and fine structures. Stable PSNR, SSIM, and LPIPS at the projection level across views confirm prediction of an unobserved spectral channel without direct supervision from that channel.

\section{Conclusion}

We proposed 4D-SG for sparse-view spectral CT reconstruction, which learned shared Gaussian geometry to represent common 3D structure and employed GSC-Net to predict raw density transformation curves for each Gaussian to model attenuation variation across spectral channels. Experiments on synthesized datasets, datasets with simulated projections, and real projection datasets demonstrated improved reconstruction quality, spectral consistency across channels, and computational efficiency over conventional baselines based on voxel grids and Gaussian representations. Beyond sparse-view reconstruction, the learned shared-structure 4D spectral Gaussian representation may also provide a useful basis for downstream spectral CT applications, including material decomposition, virtual monoenergetic imaging, and other spectral reconstruction and analysis tasks, which will be explored in future work.

\bibliographystyle{IEEEtran}
\bibliography{references}

@article{mccollough2015dual,
  title={Dual-and multi-energy {CT}: principles, technical approaches, and clinical applications},
  author={McCollough, Cynthia H and Leng, Shuai and Yu, Lifeng and Fletcher, Joel G},
  journal={Radiology},
  volume={276},
  number={3},
  pages={637--653},
  year={2015},
  publisher={Radiological Society of North America}
}

@article{willemink2018photon,
  title={Photon-counting {CT}: technical principles and clinical prospects},
  author={Willemink, Martin J and Persson, Mats and Pourmorteza, Amir and Pelc, Norbert J and Fleischmann, Dominik},
  journal={Radiology},
  volume={289},
  number={2},
  pages={293--312},
  year={2018},
  publisher={Radiological Society of North America}
}

@article{greffier2023spectral,
  title={Spectral {CT} imaging: technical principles of dual-energy {CT} and multi-energy photon-counting {CT}},
  author={Greffier, Jo{\"e}l and Villani, Nicolas and Defez, Didier and Dabli, Djamel and Si-Mohamed, Salim},
  journal={Diagnostic and Interventional Imaging},
  volume={104},
  number={4},
  pages={167--177},
  year={2023},
  publisher={Elsevier}
}

@article{wang2021spectralpcct,
  title={Spectral photon counting {CT}: Imaging algorithms and performance assessment},
  author={Wang, Adam S and Pelc, Norbert J},
  journal={IEEE Transactions on Radiation and Plasma Medical Sciences},
  volume={5},
  number={4},
  pages={453--464},
  year={2020},
  publisher={IEEE}
}

@article{bousse2024review,
  title={Systematic review on learning-based spectral {CT}},
  author={Bousse, Alexandre and Kandarpa, Venkata Sai Sundar and Rit, Simon and Perelli, Alessandro and Li, Mengzhou and Wang, Guobao and Zhou, Jian and Wang, Ge},
  journal={IEEE Transactions on Radiation and Plasma Medical Sciences},
  volume={8},
  number={2},
  pages={113--137},
  year={2023},
  publisher={IEEE}
}

@article{long2014multimaterial,
  title={Multi-material decomposition using statistical image reconstruction for spectral {CT}},
  author={Long, Yong and Fessler, Jeffrey A},
  journal={IEEE Transactions on Medical Imaging},
  volume={33},
  number={8},
  pages={1614--1626},
  year={2014},
  publisher={IEEE}
}

@article{rigie2015joint,
  title={Joint reconstruction of multi-channel, spectral {CT} data via constrained total nuclear variation minimization},
  author={Rigie, David S and La Riviere, Patrick J},
  journal={Physics in Medicine \& Biology},
  volume={60},
  number={5},
  pages={1741--1762},
  year={2015},
  publisher={IOP Publishing}
}

@article{liu2016ticmr,
  title={{TICMR}: Total image constrained material reconstruction via nonlocal total variation regularization for spectral {CT}},
  author={Liu, Jiulong and Ding, Huanjun and Molloi, Sabee and Zhang, Xiaoqun and Gao, Hao},
  journal={IEEE Transactions on Medical Imaging},
  volume={35},
  number={12},
  pages={2578--2586},
  year={2016},
  publisher={IEEE}
}

@article{zhang2017tensor,
  title={Tensor-based dictionary learning for spectral {CT} reconstruction},
  author={Zhang, Yanbo and Mou, Xuanqin and Wang, Ge and Yu, Hengyong},
  journal={IEEE Transactions on Medical Imaging},
  volume={36},
  number={1},
  pages={142--154},
  year={2016},
  publisher={IEEE}
}

@article{schmidt2017basis,
  title={A spectral {CT} method to directly estimate basis material maps from experimental photon-counting data},
  author={Schmidt, Taly Gilat and Barber, Rina Foygel and Sidky, Emil Y},
  journal={IEEE Transactions on Medical Imaging},
  volume={36},
  number={9},
  pages={1808--1819},
  year={2017},
  publisher={IEEE}
}

@article{mechlem2018joint,
  title={Joint statistical iterative material image reconstruction for spectral computed tomography using a semi-empirical forward model},
  author={Mechlem, Korbinian and Ehn, Sebastian and Sellerer, Thorsten and Braig, Eva and M{\"u}nzel, Daniela and Pfeiffer, Franz and No{\"e}l, Peter B},
  journal={IEEE Transactions on Medical Imaging},
  volume={37},
  number={1},
  pages={68--80},
  year={2017},
  publisher={IEEE}
}

@article{shen2023josr,
  title={Joint reconstruction and spectrum refinement for photon-counting-detector spectral {CT}},
  author={Shen, Le and Xing, Yuxiang and Zhang, Li},
  journal={IEEE Transactions on Medical Imaging},
  volume={42},
  number={9},
  pages={2653--2665},
  year={2023},
  publisher={IEEE}
}

@article{hu2024prospect,
  title={Spectral {CT} image reconstruction using a constrained optimization approach---An algorithm for {AAPM} 2022 spectral {CT} grand challenge and beyond},
  author={Hu, Xiaoyu and Jia, Xun},
  journal={Medical Physics},
  volume={51},
  number={5},
  pages={3376--3390},
  year={2024},
  publisher={Wiley Online Library}
}

@article{sidky2024challenge,
  title={Report on the AAPM deep-learning spectral {CT} Grand Challenge},
  author={Sidky, Emil Y and Pan, Xiaochuan},
  journal={Medical Physics},
  volume={51},
  number={2},
  pages={772--785},
  year={2024},
  publisher={Wiley Online Library}
}

@article{wu2021dlspectral,
  title={Deep learning based spectral {CT} imaging},
  author={Wu, Weiwen and Hu, Dianlin and Niu, Chuang and Vanden Broeke, Lieza and Butler, Anthony PH and Cao, Peng and Atlas, James and Chernoglazov, Alexander and Vardhanabhuti, Varut and Wang, Ge},
  journal={Neural Networks},
  volume={144},
  pages={342--358},
  year={2021},
  publisher={Elsevier}
}

@article{chen2024soulnet,
  title={{SOUL-net}: A sparse and low-rank unrolling network for spectral {CT} image reconstruction},
  author={Chen, Xiang and Xia, Wenjun and Yang, Ziyuan and Chen, Hu and Liu, Yan and Zhou, Jiliu and Wang, Zhe and Chen, Yang and Wen, Bihan and Zhang, Yi},
  journal={IEEE Transactions on Neural Networks and Learning Systems},
  volume={35},
  number={12},
  pages={18620--18634},
  year={2023},
  publisher={IEEE}
}

@article{shi2024textureprior,
  title={Learned tensor neural network texture prior for photon-counting {CT} reconstruction},
  author={Shi, Yongyi and Gao, Yongfeng and Xu, Qiong and Li, Yang and Mou, Xuanqin and Liang, Zhengrong},
  journal={IEEE Transactions on Medical Imaging},
  volume={43},
  number={11},
  pages={3830--3842},
  year={2024},
  publisher={IEEE}
}

@article{ge2023mbdectnet,
  title={{MB-DECTNet}: a model-based unrolling network for accurate {3D} dual-energy {CT} reconstruction from clinically acquired helical scans},
  author={Ge, Tao and Liao, Rui and Medrano, Maria and Politte, David G and Williamson, Jeffrey F and O'Sullivan, Joseph A},
  journal={Physics in Medicine \& Biology},
  volume={68},
  number={24},
  pages={245009},
  year={2023},
  publisher={IOP Publishing}
}

@article{liu2024sgm,
  title={Sparse-view spectral {CT} reconstruction and material decomposition based on multi-channel {SGM}},
  author={Liu, Yuedong and Zhou, Xuan and Wei, Cunfeng and Xu, Qiong},
  journal={IEEE Transactions on Medical Imaging},
  volume={43},
  number={10},
  pages={3425--3435},
  year={2024},
  publisher={IEEE}
}

@article{liu2025spn,
  title={Low-dose spectral {CT} reconstruction based on structural prior network},
  author={Liu, Yuedong and Zhou, Xuan and Wang, Chengmin and Wei, Cunfeng and Xu, Qiong},
  journal={Medical Physics},
  volume={52},
  number={10},
  pages={e70061},
  year={2025},
  publisher={Wiley Online Library}
}

@article{guo2025coupled,
  title={Sparse-view spectral {CT} reconstruction via a coupled subspace representation and score-based generative model},
  author={Guo, Jie and Wang, Yizhong and Wang, Shaoyu and Zheng, Zhizhong and Li, Lei and Cai, Ailong and Yan, Bin},
  journal={Quantitative Imaging in Medicine and Surgery},
  volume={15},
  number={6},
  pages={5474--5495},
  year={2025},
  publisher={LWW}
}

@article{zhang2024oign,
  title={One-step inverse generation network for sparse-view dual-energy {CT} reconstruction and material imaging},
  author={Zhang, Xinrui and Li, Lei and Wang, Shaoyu and Liang, Ningning and Cai, Ailong and Yan, Bin},
  journal={Physics in Medicine \& Biology},
  volume={69},
  number={14},
  pages={145012},
  year={2024},
  publisher={IOP Publishing}
}

@inproceedings{song2023piner,
  title={{PINER}: Prior-informed implicit neural representation learning for test-time adaptation in sparse-view {CT} reconstruction},
  author={Song, Bowen and Shen, Liyue and Xing, Lei},
  booktitle={Proceedings of the IEEE/CVF Winter Conference on Applications of Computer Vision},
  pages={1928--1938},
  year={2023}
}

@inproceedings{lin2023difnet,
  title={Learning deep intensity field for extremely sparse-view {CBCT} reconstruction},
  author={Lin, Yiqun and Luo, Zhongjin and Zhao, Wei and Li, Xiaomeng},
  booktitle={International Conference on Medical Image Computing and Computer-Assisted Intervention},
  pages={13--23},
  year={2023},
  organization={Springer}
}

@inproceedings{lin2024c2rv,
  author={Lin, Yiqun and Yang, Jiewen and Wang, Hualiang and Ding, Xinpeng and Zhao, Wei and Li, Xiaomeng},
  title={{C$^2$RV}: Cross-Regional and Cross-View Learning for Sparse-View {CBCT} Reconstruction},
  booktitle={Proceedings of the IEEE/CVF Conference on Computer Vision and Pattern Recognition (CVPR)},
  month={June},
  year={2024},
  pages={11205--11214}
}

@inproceedings{cai2024saxnerf,
  title={Structure-aware sparse-view {X-ray} {3D} reconstruction},
  author={Cai, Yuanhao and Wang, Jiahao and Yuille, Alan and Zhou, Zongwei and Wang, Angtian},
  booktitle={Proceedings of the IEEE/CVF Conference on Computer Vision and Pattern Recognition},
  pages={11174--11183},
  year={2024}
}

@article{kerbl2023gaussian,
  title={{3D} {Gaussian} splatting for real-time radiance field rendering},
  author={Kerbl, Bernhard and Kopanas, Georgios and Leimk{\"u}hler, Thomas and Drettakis, George and others},
  journal={ACM Trans. Graph.},
  volume={42},
  number={4},
  pages={139--1},
  year={2023}
}

@inproceedings{cai2024xgaussian,
  title={Radiative {Gaussian} splatting for efficient {X-ray} novel view synthesis},
  author={Cai, Yuanhao and Liang, Yixun and Wang, Jiahao and Wang, Angtian and Zhang, Yulun and Yang, Xiaokang and Zhou, Zongwei and Yuille, Alan},
  booktitle={European Conference on Computer Vision},
  pages={283--299},
  year={2024},
  organization={Springer}
}

@article{zha2024r2gaussian,
  title={{R$^2$-Gaussian}: Rectifying Radiative Gaussian Splatting for Tomographic Reconstruction},
  author={Zha, Ruyi and Lin, Tao Jun and Cai, Yuanhao and Cao, Jiwen and Zhang, Yanhao and Li, Hongdong},
  journal={arXiv preprint arXiv:2405.20693},
  year={2024}
}

@inproceedings{lin2024learning,
  title={Learning {3D} {Gaussians} for extremely sparse-view cone-beam {CT} reconstruction},
  author={Lin, Yiqun and Wang, Hualiang and Chen, Jixiang and Li, Xiaomeng},
  booktitle={International Conference on Medical Image Computing and Computer-Assisted Intervention},
  pages={425--435},
  year={2024},
  organization={Springer}
}

@article{li20253dgrct,
  title={{3DGR-CT}: Sparse-view {CT} reconstruction with a {3D} Gaussian representation},
  author={Li, Yingtai and Fu, Xueming and Li, Han and Zhao, Shang and Jin, Ruiyang and Zhou, S Kevin},
  journal={Medical Image Analysis},
  volume={103},
  pages={103585},
  year={2025},
  publisher={Elsevier}
}

@misc{aapm2017lowdose,
  author={{American Association of Physicists in Medicine}},
  title={{Low dose CT grand challenge}},
  year={2017},
  howpublished={[Online]},
  url={http://www.aapm.org/GrandChallenge/LowDoseCT/},
  note={Accessed: Apr. 6, 2017}
}

@misc{klacansky2022scientific,
  author={Klacansky, Pavol},
  title={{Scientific Visualization Datasets}},
  year={2022},
  howpublished={[Online]},
  url={https://klacansky.com/open-scivis-datasets/}
}

@article{zhou2025cone,
  title={A cone-beam photon-counting {CT} dataset for spectral image reconstruction and deep learning},
  author={Zhou, Enze and Li, Wenjian and Xu, Wenting and Wan, Kefei and Lu, Yuwei and Chen, Shangbin and Zheng, Gang and Xie, Tianwu and Liu, Qian},
  journal={Scientific Data},
  volume={12},
  number={1},
  pages={1955},
  year={2025},
  publisher={Nature Publishing Group UK London}
}

@misc{volgyes2018dual,
  author={V{\"o}lgyes, David},
  title={Dual Energy {CT} Scan of Ordinary Objects},
  year={2018},
  month={jun},
  day={26},
  publisher={Zenodo},
  doi={10.5281/zenodo.1253035},
  url={https://doi.org/10.5281/zenodo.1253035}
}

@inproceedings{zha2022naf,
  title={{NAF}: Neural Attenuation Fields for Sparse-View {CBCT} Reconstruction},
  author={Zha, Ruyi and Zhang, Yanhao and Li, Hongdong},
  booktitle={International Conference on Medical Image Computing and Computer-Assisted Intervention},
  pages={442--452},
  year={2022},
  organization={Springer}
}

\end{document}